\documentclass{article}
\PassOptionsToPackage{numbers,sort&compress}{natbib}
\usepackage[T1]{fontenc} 
\usepackage{iclr2027_conference,times}
\setcitestyle{numbers,square,comma,sort&compress}
\usepackage{graphicx}
\usepackage{xcolor}
\usepackage{amssymb}
\usepackage{booktabs,tabularx,array,longtable}
\usepackage{colortbl}
\definecolor{baserow}{gray}{0.92}
\definecolor{oursrow}{RGB}{226,236,252}
\usepackage{float}
\usepackage{fvextra}
\usepackage{placeins}
\usepackage{hyperref}
\usepackage{url}
\graphicspath{{figures/}}
\newcolumntype{Y}{>{\raggedright\arraybackslash}X}
\newcommand{\paperimage}[2][\linewidth]{\includegraphics[width=#1,height=0.36\textheight,keepaspectratio]{#2}}
\title{EgoOmniWorld: Unifying Egocentric Understanding via Ego–Action–World Interactions}
\iclrfinalcopy

\author{{\normalsize\normalfont
\textbf{Songruo Chen}$^{1}$\thanks{Equal contribution.} \quad
\textbf{Yuqian Fu}$^{2*}$\thanks{Corresponding author.} \quad
\textbf{Yu Li}$^{1*}$ \quad
\textbf{Mingkang Dong}$^{3}$ \quad
\textbf{Sifan Zhou}$^{4}$} \\
{\normalsize\normalfont
\textbf{Bin Ren}$^{5}$ \quad
\textbf{Xu Zheng}$^{6}$ \quad
\textbf{Tianwen Qian}$^{7}$ \quad
\textbf{Mohamed Elhoseiny}$^{2}$ \quad
\textbf{Yanwei Fu}$^{1}$} \\[0.5ex]
{\small\normalfont
$^{1}$Fudan University \quad
$^{2}$KAUST \quad
$^{3}$Universiti Malaya \quad
$^{4}$CMU \quad
$^{5}$MBZUAI \quad
$^{6}$Great Bay University \quad
$^{7}$ECNU}
}

\begin{document}
\raggedbottom
\maketitle
\lhead{}

\begin{figure}[!h]
\centering
\vspace{-0.15in}
\includegraphics[width=0.95\linewidth]{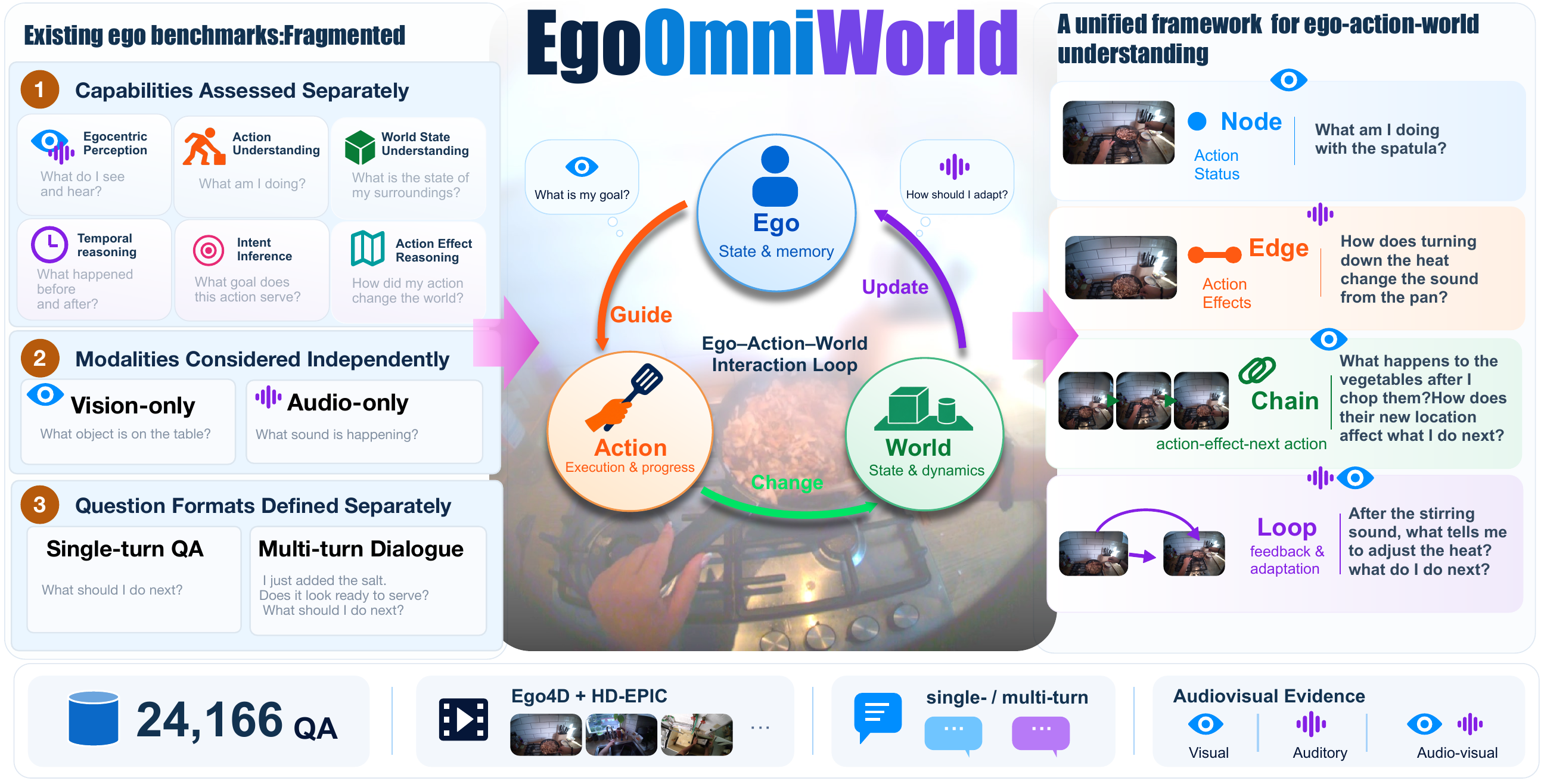}
\vspace{-0.15in}
\caption{EgoOmniWorld unifies fragmented egocentric evaluation around the Ego--Action--World
interaction loop, organizing questions into Node, Edge, Chain, and Loop structures.}
\label{fig:teaser}
\end{figure}

\begin{abstract}

Egocentric video understanding has rapidly expanded into a diverse landscape of benchmarks targeting different capabilities, modalities, and evaluation formats. While each direction captures an important aspect of egocentric understanding, these capabilities are often studied in isolation, despite being tightly coupled in real-world interaction. We argue that a defining property of egocentric video provides a natural unifying principle: \textit{the wearer perceives the world, forms goals, takes actions, changes the world, and observes the resulting feedback to guide subsequent behavior, forming a closed \textbf{Ego--Action--World} interaction loop}.
Based on this view, we introduce \textbf{EgoOmniWorld}, a unified benchmark that organizes egocentric understanding through four complementary interaction structures: \textit{Node}, capturing ego, action, and world states; \textit{Edge}, capturing directed relations among them; \textit{Chain}, capturing linked dependencies across multiple relations; and \textit{Loop}, capturing feedback-closed interactions.
We further develop an \textit{interaction-first, evidence-grounded data pipeline} that extracts structured interactions from egocentric videos, grounds them in visual and auditory evidence, and generates questions according to their interaction structure. Rather than treating multimodality and dialogue form as separate task families, the pipeline naturally determines the required modality and question-answering format according to the evidence and dependencies involved.
EgoOmniWorld contains 24,166 QA groups, including 20,284 training and 3,882 test samples, spanning diverse interaction scopes, modality requirements, and single- and multi-turn settings. Experiments on representative open- and closed-source multimodal large language models (MLLMs) reveal substantial and uneven limitations across Ego--Action--World interactions. Models trained on EgoOmniWorld consistently improve in-domain interaction understanding and generalize to existing egocentric benchmarks.
We hope EgoOmniWorld provides a unified foundation for systematically evaluating and advancing egocentric video understanding. Data, code, and models will be released.

\end{abstract}
\section{Introduction}
Egocentric video understanding has rapidly evolved into a broad and increasingly diverse research landscape~\cite{zhang2023egospot, mahdi2025exo2egosyn, fu2025objectrelator, pan2026v}. Existing benchmarks target a wide range of capabilities, including visual perception~\cite{mangalam2023egoschema, li2026egocross}, action understanding~\cite{jia2022egotaskqa, rodin2025easg}, spatial and temporal reasoning~\cite{plizzari2025omnia, zhang2026egonight, xu2025tog}, intent inference~\cite{li2025egotom, pan2026egointent}, and planning prediction~\cite{cheng2024videgothink,chen2026egoplan}. Recent efforts further span different modality requirements, from vision-only to joint audio-visual understanding~\cite{zhu2026egosound, seth2026egoavu}, and evaluation formats, from single-turn question answering to multi-turn dialogue~\cite{ye2025mmego, gao2026lifeeval}. While each direction captures an important aspect of egocentric understanding, the resulting benchmark landscape remains fragmented, with different capabilities, modalities, and evaluation formats often studied under separate task definitions and protocols. The field therefore lacks a common framework for systematically organizing these diverse aspects and understanding how they relate to one another.

A defining characteristic of egocentric video is that it records the world from the perspective of an active participant rather than a passive observer. The wearer continuously perceives the surrounding world, forms goals, and takes actions to achieve them, while the resulting changes in the world provide feedback that guides subsequent decisions and behavior. We refer to this closed process as the \textbf{Ego--Action--World interaction loop}, which naturally connects perception, action understanding, intent reasoning, world-state change, and feedback within a unified framework.

Based on this view, we introduce \textbf{EgoOmniWorld}, a unified benchmark for egocentric video understanding centered on Ego--Action--World interactions. Rather than organizing questions around isolated capability labels, EgoOmniWorld characterizes complementary structural scopes of the interaction process. \textit{Node} captures individual Ego, Action, and World states; \textit{Edge} captures directed relations among the nodes; \textit{Chain} captures linked dependencies across multiple relations; and \textit{Loop} captures feedback-closed interactions in which action outcomes are perceived by the ego and influence subsequent goals or behavior. Note that these structures represent different scopes of interaction composition rather than a hierarchy of task difficulty. 
Likewise, we argue that modality and dialogue form should also not be treated as independent task dimensions. Instead, they arise naturally from the Ego--Action--World interaction process itself: different stages of the interaction call for different evidence, whether visual or audio-visual, and different dependency structures are naturally expressed through single- or multi-turn QA.
In this way, EgoOmniWorld unifies not only diverse understanding capabilities, but also the modalities and dialogue formats through which they are most naturally evaluated.

To instantiate this framework, we develop an \textit{interaction-first, evidence-grounded data engine} over Ego4D~\cite{grauman2022ego4d} and HD-EPIC~\cite{perrett2025hd}. As illustrated in Figure~\ref{fig:teaser}, rather than generating isolated questions first and composing them afterward, the data engine first reconstructs structured Ego--Action--World interactions from egocentric videos. It identifies ego states and goals, executed actions, world states and changes, and observable feedback; grounds these components in visual and auditory evidence; and aggregates related interactions through persistent entities and task episodes. Questions are generated only after the target interaction structure and its supporting evidence have been established, with modality requirements and dialogue forms determined by the evidence and dependencies involved.
This interaction-first design enables Node and Edge questions to be derived from local interaction components, Chain questions from linked dependencies, and Loop questions from feedback-closed cycles, while preserving traceable evidence for each answer. Generated candidates further undergo shared quality checks, structure-specific validation, constrained repair, and revalidation. In total, the resulting EgoOmniWorld benchmark contains 24,166 evidence-grounded QA groups, including 20,284 for training and 3,882 for testing.

Extensive experiments on various open- and closed-source multimodal large language models (MLLMs) reveal three key findings. First, current models exhibit markedly different strengths across Ego--Action--World interaction structures and semantic categories, suggesting that aggregate performance alone obscures important capability gaps. Second, training on EgoOmniWorld consistently improves in-domain interaction understanding under both LoRA and full fine-tuning; for example, Qwen2.5-Omni-7B improves from 36.42\% to 47.36\% with LoRA and 48.34\% with full fine-tuning. 
Third, the trained model demonstrates zero-shot transfer to EgoSound, EgoTempo, and EgoToM, achieving gains of up to 8.08 percentage points.

In summary, our contributions are threefold: 
1) we introduce an \textbf{Ego--Action--World interaction framework} with four complementary structures---Node, Edge, Chain, and Loop---for systematically capturing diverse aspects of egocentric understanding; 
2) we instantiate this framework as \textbf{EgoOmniWorld}, a 24,166-sample evidence-grounded benchmark constructed with an interaction-first data engine, spanning diverse capability scopes, modality requirements, and evaluation formats; and 
3) we demonstrate that EgoOmniWorld serves as both a unified diagnostic testbed and a source of transferable supervision for egocentric video understanding.

\section{Related Work}
\label{sec:related}

\noindent\textbf{Egocentric Video Benchmarks.}
A growing body of egocentric video benchmarks has explored increasingly diverse aspects of first-person understanding. 
EgoSchema~\cite{mangalam2023egoschema} and EgoTempo~\cite{plizzari2025egotempo} emphasize long-horizon and temporal reasoning; 
EgoTaskQA~\cite{jia2022egotaskqa} studies task-oriented understanding, including action dependencies and goal-related reasoning; 
EgoToM~\cite{li2025egotom} focuses on reasoning about the wearer's mental states, goals, and actions; 
and MultiHop-EgoQA~\cite{chen2025grounded} targets multi-hop reasoning across temporally distributed evidence.
Other benchmarks, e.g., EgoCross~\cite{li2026egocross}, EgoNight~\cite{zhang2026egonight}, and EgoTextVQA~\cite{zhou2025egotextvqa} have explored complementary aspects such as cross-domain generalization, nighttime understanding, and scene-text understanding.

Beyond capability-specific evaluation, recent benchmarks have expanded along modality and dialogue dimensions.
EgoLifeQA~\cite{yang2025egolife}, EgoSound~\cite{zhu2026egosound}, and EgoAVU~\cite{seth2026egoavu} extend egocentric understanding from vision-only input to audio-visual reasoning. Meanwhile, recent efforts e.g., LifeEval~\cite{gao2026lifeeval} move beyond single-turn QA toward multi-turn or dialogue-based evaluation.
Table~\ref{tab:benchmark_structure} provides a representative, rather than exhaustive, comparison of existing benchmarks in terms of interaction structure, input modality, and dialogue format. Viewed through the Ego--Action--World lens, these works substantially broaden the landscape of egocentric understanding, but remain organized around different task definitions, modality settings, and evaluation protocols, without a common structure connecting them. EgoOmniWorld instead treats these dimensions as natural properties of the underlying interaction: the required modality follows from the available supporting evidence, while single- or multi-turn QA follows from the dependency structure being evaluated.

\begin{table}[!htbp]
\centering
\caption{
Representative coverage of existing egocentric benchmarks under the EgoOmniWorld taxonomy.
E, A, and W denote Ego, Action, and World; arrows indicate Edge directions.
$\checkmark$: explicit coverage; $\sim$: related but not explicitly structured. 
}

\label{tab:benchmark_structure}

\begingroup
\footnotesize
\setlength{\tabcolsep}{3.5pt}
\renewcommand{\arraystretch}{1.12}

\begin{tabularx}{\linewidth}{@{}l c c Y c c c@{}}
\toprule
\textbf{Benchmark} & \textbf{Input} & \textbf{Node} & \textbf{Edge} & \textbf{Chain} & \textbf{Loop} 
& \textbf{Dialogue} \\
\midrule

EgoSchema
& Video
& A, W
& A$\rightarrow$E
& $\sim$
& --
& single-turn \\

EgoTempo
& Video
& A, W
& E$\rightarrow$A
& $\sim$
& --
& single-turn \\

EgoTaskQA
& Video
& E, A, W
& A$\rightarrow$E; A$\rightarrow$W; W$\rightarrow$A
& $\sim$
& --
& single-turn \\

EgoToM
& Video
& E, A
& A$\rightarrow$E
& $\sim$
& --
& single-turn \\

MultiHop-EgoQA
& Video
& A, W
& --
& $\sim$
& --
& single-turn \\

EgoLifeQA
& Video + Audio
& E, A, W
& W$\rightarrow$E ; E$\rightarrow$A
& $\sim$
& --
& single-turn \\

EgoSound
& Video + Audio
& E, A, W
&  E$\rightarrow$W
& --
& --
& single-turn \\

LifeEval
& Video
& E, A, W
& A$\leftrightarrow$E ; W$\rightarrow$A
& $\sim$
& --
& single- \& multi-turn \\

\midrule
\rowcolor{oursrow}
\textbf{EgoOmniWorld}
& \textbf{Video + Audio}
& \textbf{E, A, W}
& \textbf{E$\leftrightarrow$A; E$\leftrightarrow$W; A$\leftrightarrow$W}
& \textbf{\checkmark} 
& \textbf{\checkmark}
& \textbf{Single- \& Multi-turn} \\
\bottomrule
\end{tabularx}
\endgroup
\end{table}

\noindent\textbf{Omni-modal Multimodal LLMs.}
Recent multimodal large language models (MLLMs) increasingly support joint video, audio, and language understanding. Representative systems include proprietary models such as Gemini~\cite{team2023gemini} and Qwen-Omni-Plus/Flash~\cite{team2026qwen3}, as well as open-source models including MiniCPM-o~\cite{cui2026minicpm}, video-SALMONN~\cite{sun2025video}, VideoLLaMA2~\cite{cheng2024videollama}, Qwen2.5-Omni~\cite{Xu2025Qwen25OmniTR}, and Qwen3-Omni~\cite{team2026qwen3}.
These models provide a strong foundation for evaluating holistic egocentric understanding across heterogeneous visual, auditory, and interaction cues. We therefore benchmark representative closed- and open-source models from these families on EgoOmniWorld, and further study whether EgoOmniWorld supervision improves interaction understanding and directly transfers to existing egocentric benchmarks.

\section{EgoOmniWorld Benchmark Construction}

{This section presents the core design of EgoOmniWorld (Section~\ref{sec:taxonomy}), its interaction-first data engine for structuring Ego--Action--World interactions into evidence-grounded VQA (Section~\ref{sec:construction}), and the resulting benchmark statistics (Section~\ref{sec:dataset_statistics}). 

\subsection{Core Design Principles}
\label{sec:taxonomy}

\begin{figure}[t]
\centering
\includegraphics[width=0.95\linewidth,height=0.24\textheight,keepaspectratio]{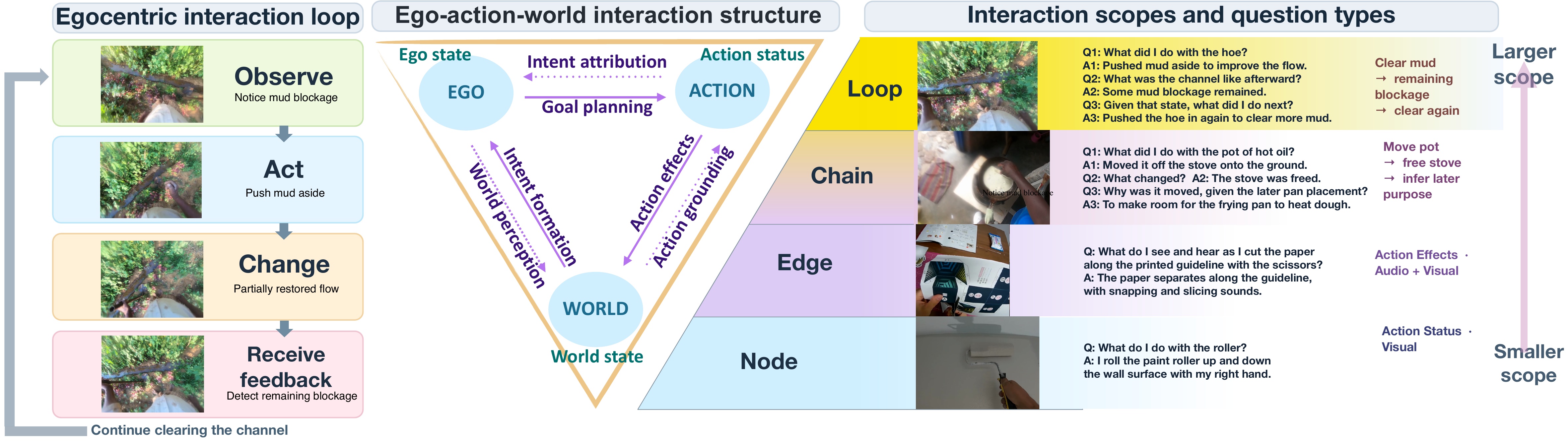}
\caption{
The EgoOmniWorld taxonomy. Left: an egocentric interaction loop. Middle: the Ego--Action--World structure, comprising three node states and six directed edge relations. Right: four complementary interaction structures, from Node to Loop, with example questions.
}
\label{fig:taxonomy}
\end{figure}

EgoOmniWorld is built around the defining closed-loop nature of egocentric interaction: the wearer perceives the world, acts according to a goal, changes the world, and observes the resulting state to guide subsequent behavior.
We organize this process around three design principles.

\noindent\textbf{Ego--Action--World interactions are the unit of organization.}
Each interaction is represented by three components.
The \emph{Ego} captures the wearer's interaction-relevant state, including observations, goals, and current subgoals;
the \emph{Action} captures executed behavior and its progress;
and the \emph{World} captures external entities, their states, and state changes.
Together, these components provide a common representation for organizing diverse egocentric understanding capabilities.
Related interactions can further be grouped into task episodes to capture dependencies beyond a single event.

\noindent\textbf{Four structures capture complementary interaction scopes.}
Rather than defining tasks by isolated capability labels, EgoOmniWorld derives four structural scopes from the Ego--Action--World process, as in Figure~\ref{fig:taxonomy}.
\emph{Node} targets individual Ego, Action, or World states, corresponding to \emph{Ego State}, \emph{Action Status}, and \emph{World State}.
\emph{Edge} targets one directed relation between two components: Ego$\to$Action (\emph{Goal Planning}), Action$\to$Ego (\emph{Intent Attribution}), Action$\to$World (\emph{Action Effects}), World$\to$Action (\emph{Action Grounding}), World$\to$Ego (\emph{Intent Formation}), and Ego$\to$World (\emph{World Perception}).
\emph{Chain} composes multiple directed relations into linked dependencies, where earlier states, actions, or observations are required to understand later events.
\emph{Loop} closes the interaction cycle by capturing how action outcomes are perceived by the ego and subsequently confirm, adjust, or change its goals or behavior.
Thus, Node, Edge, Chain, and Loop correspond respectively to interaction elements, directed relations, their composition, and feedback closure, rather than a hierarchy of task difficulty.

\noindent\textbf{Modality and dialogue form arise from the interaction.}
Visual and auditory signals are treated as complementary evidence for Ego--Action--World interactions rather than as separate task families.
The modality requirement of each question is determined by the evidence necessary to answer it, yielding visual, auditory, or audio-visual questions.
Likewise, dialogue form follows the dependency structure being evaluated: local states or relations can be probed in a single turn, while multi-turn QA can expose intermediate states and dependencies across an interaction.
Modality and dialogue form are therefore orthogonal to the Node--Edge--Chain--Loop structure and are instantiated according to the underlying evidence and interaction dependencies.

\subsection{Data Engine Pipeline}
\label{sec:construction}
To instantiate the Ego--Action--World framework, we develop an
interaction-first, evidence-grounded data engine that reconstructs
structured interactions before generating VQA. We select activity-diverse
source videos from Ego4D~\cite{grauman2022ego4d} and HD-EPIC~\cite{perrett2025hd}, retaining their audio, narrations,
and timestamps. Narration-based clustering promotes activity diversity
in Ego4D, while official HD-EPIC annotations help identify candidate
video segments. Narrations serve as temporal anchors rather than
substitutes for evidence from the original videos.

As illustrated in Figure~\ref{fig:pipeline}, the data engine comprises
four stages: 1) local interaction extraction, 2) multimodal evidence grounding,
3) episode aggregation and dependency linking, and 4) evidence-grounded QA
generation and validation. 

Source references and timestamps are
preserved throughout the pipeline, enabling each generated question
to be traced back to its supporting observations. Detailed construction
rules and prompt templates are provided in Appendix~\ref{app:prompts}.

\begin{figure}[t]
\centering
\paperimage{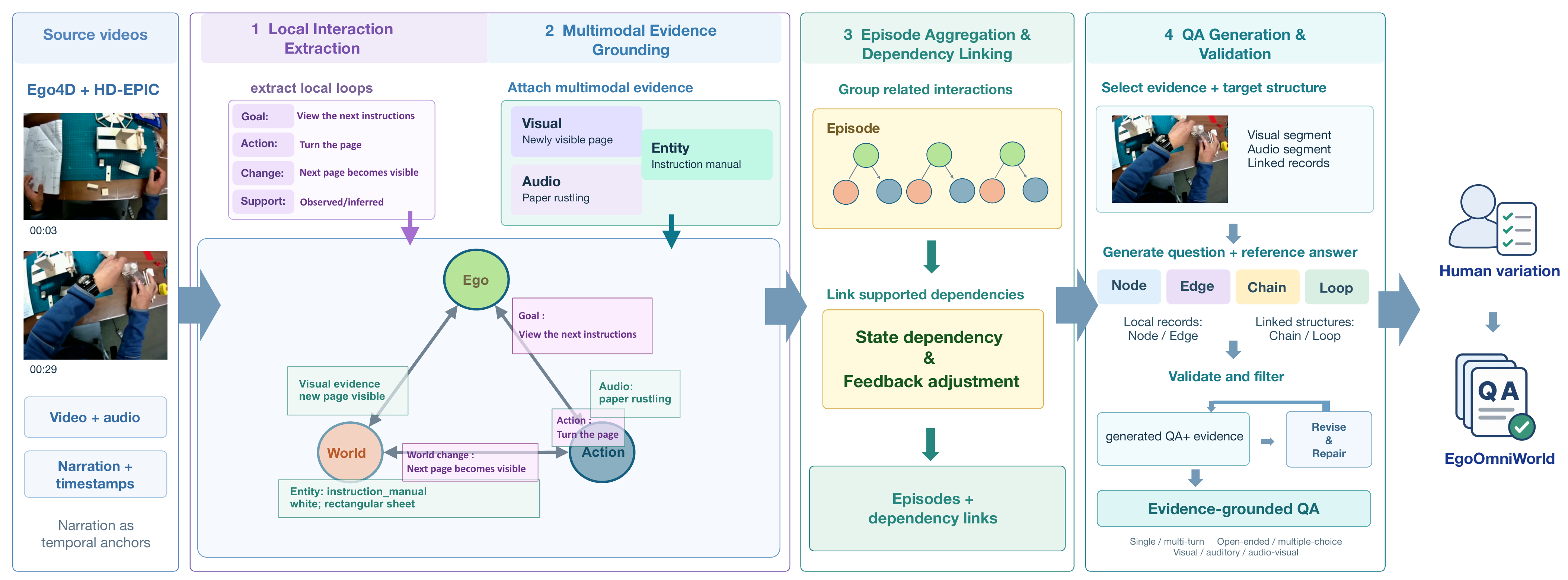}
\caption{
Overview of the proposed data engine.
Given source egocentric videos, the four stages extract local
Ego--Action--World interactions, ground them in multimodal evidence,
aggregate episodes and link dependencies, and generate and validate
evidence-grounded QA. 
}
\label{fig:pipeline}
\end{figure}

\noindent\textbf{Step 1: Local Interaction Extraction.}
We first segment each video window into individual actions. For each
action, we record the world state before it, the goal and current subgoal it
serves, its outcome, and the feedback received by the wearer. A goal is
recorded only when it is supported by speech, context, or later consequences.
Failed and interrupted actions are retained, as they often reveal how feedback
changes behavior. The output is a set of timestamped interaction records.

\noindent\textbf{Step 2: Multimodal Evidence Grounding.}
We then attach supporting evidence to each interaction record. Auditory
evidence includes speech and action-related sounds, and visual evidence
includes object properties, hand--object contacts, and scene layout. Each
observation is timestamped and labeled with its modality. Entity identifiers
link repeated observations of the same object across windows, while its states
at different times are kept separate. For example, turning a page of a manual
is grounded in both the newly visible page and the paper-rustling sound. These
modality labels later determine the modality requirement of each question.

\noindent\textbf{Step 3: Episode Aggregation and Dependency Linking.}
Chain and Loop questions require connections across multiple
interactions. We therefore group adjacent interactions that serve the same
activity stage into task episodes. Two episodes are linked only when a state,
instruction, failure, or unfinished task carries over from one to the other.
Temporal adjacency or a shared object alone is not sufficient. Each link
records the connected components, the transferred state, and the required
modalities. The output is a set of episodes and dependency links, from which
Chain and Loop questions are generated.

\noindent\textbf{Step 4: Evidence-Grounded QA Generation and Validation.}
QA generation follows an evidence-first principle: the supporting
evidence and the target structure are selected before the question and answer
are written. Node and Edge questions are generated from local interaction
records, and Chain and Loop questions from linked episodes. In multi-turn QA,
the final turn must combine earlier turns with new evidence, so it cannot be
answered from the dialogue history alone. All candidates are then checked for
evidential support, temporal consistency, modality correctness, answer
leakage, and structural validity. Failed candidates are repaired using existing
records only and then re-validated. A random subset of the final QA is further
reviewed by humans (Appendix~\ref{app:human_review}).

\subsection{Dataset Statistics}
\label{sec:dataset_statistics}

Figure~\ref{fig:dataset_sta} summarizes the statistics of EgoOmniWorld across interaction structures, semantic categories, modality requirements, and dialogue formats.

\noindent\textbf{Dataset composition.}
EgoOmniWorld contains 24,166 QA groups, comprising 20,284 training and 3,882 test groups. Each single-turn QA or complete multi-turn dialogue is counted as one group.

\noindent\textbf{Interaction coverage.}
EgoOmniWorld covers all four interaction structures, with Node, Edge, Chain, and Loop accounting for 41.5\%, 39.4\%, 10.3\%, and 8.9\% of QA groups, respectively. The nine Node and Edge semantic categories further span diverse aspects of Ego--Action--World understanding. Action Status and World Perception are the most frequent, while Goal Planning, Intent Attribution, and Intent Formation are less represented. 

\noindent\textbf{Modality coverage.}
EgoOmniWorld naturally incorporates diverse modality requirements, with 46.9\% of QA groups requiring visual evidence only, 15.4\% requiring auditory evidence only, and 37.7\% requiring joint audio-visual evidence. 
Importantly, modality labels reflect the evidence necessary to answer each question rather than the modalities available in the source video.

\noindent\textbf{Dialogue and temporal coverage.}
The dataset includes 19,833 single-turn groups (82.1\%) and 4,333 multi-turn dialogues (17.9\%). Among multi-turn dialogues, 1,750 contain two turns, 2,559 contain three turns, and 24 contain four turns. Duration metadata are available for 10,664 unique video clips, with 30--60-second clips forming the largest group.

Visual examples spanning local states, directed relations, linked dependencies, and feedback-closed interactions are also demonstrated in Figure~\ref{fig:dataset_sta}. More examples are provided in Appendix.

\begin{figure}[t]
    \centering
    \includegraphics[width=\linewidth]{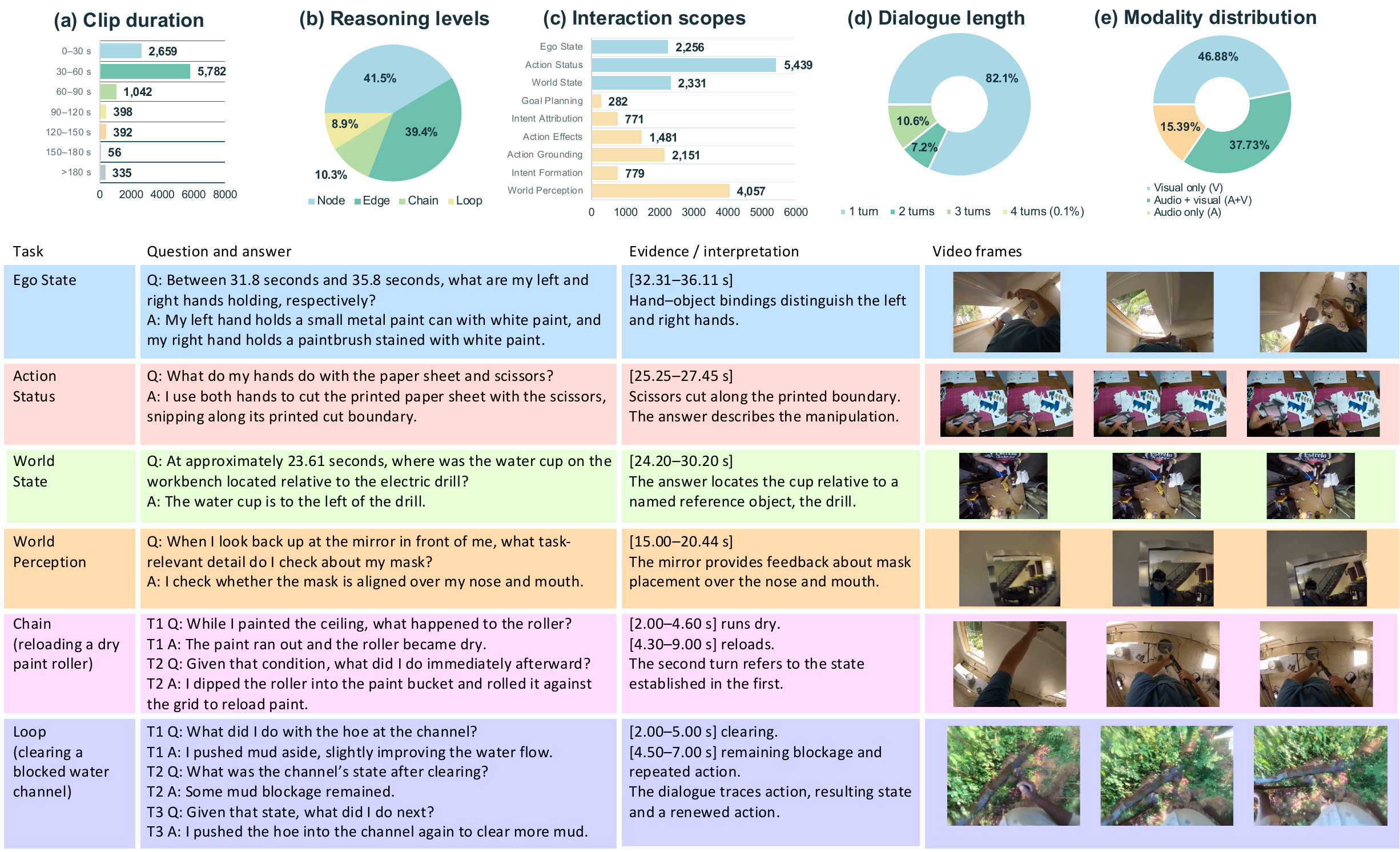}
    
    \caption{
Dataset statistics and representative QA examples in EgoOmniWorld.
Top: distributions of (a) clip duration, (b) interaction scopes,
(c) Node and Edge semantic categories, (d) dialogue length,
and (e) required evidence modality.
Bottom: representative questions, answers, supporting evidence,
and video frames illustrating individual states, directed relations,
linked dependencies, and feedback-closed interactions.
Each complete multi-turn dialogue counts as one QA group.
}

    \label{fig:dataset_sta}
\end{figure}

\section{Experiments}
EgoOmniWorld serves as a unified testbed and training resource for egocentric interaction understanding. We further investigate cross-benchmark transfer and analyze fine-grained capability gains and interaction coverage.

\subsection{Experimental Setup}
\label{sec:setup}
\noindent\textbf{Baseline models.}
We evaluate 11 open- and closed-source omni-modal MLLMs on EgoOmniWorld. The three closed-source systems include Gemini 3.8 Flash~\cite{team2023gemini}, Qwen3.5-Omni-Plus~\cite{team2026qwen3}, and Qwen3.8-Omni-Flash~\cite{team2026qwen3}. The eight open-source models include MiniCPM-o 2.6 (8B)~\cite{cui2026minicpm}, SALMONN2+ (7B and 72B)~\cite{sun2025video}, VideoLLaMA2-AV-7B~\cite{cheng2024videollama}, Qwen2.5-Omni (3B and 7B)~\cite{Xu2025Qwen25OmniTR}, and Qwen3-Omni (Thinking and Instruct variants)~\cite{team2026qwen3}. 

\noindent\textbf{EgoOmniWorld training.}
To test whether the benchmark also works as supervision, we train two open-source models, Qwen2.5-Omni-7B and Qwen3-Omni-Thinking, on the EgoOmniWorld training split. Each model is trained in two ways: LoRA, which adds low-rank adapters and leaves the base weights unchanged, and full fine-tuning (FFT), which updates the model weights directly. Training supervises the reference answers; in multi-turn dialogues, earlier turns are included and their answers are supervised as well. Videos are sampled at two frames per second with at most 64 frames and 200,704 pixels per frame, and the audio track is kept. Both regimes share the same schedule and precision settings; learning rates, adapter targets, and distributed settings are listed in Appendix~\ref{app:training_settings}.

\noindent\textbf{In-domain inference protocol.}
All models are evaluated on the EgoOmniWorld test set. For multi-turn QA, models condition on their own previous responses rather than reference answers, thereby evaluating their ability to maintain consistency across dialogue turns. Predictive questions are restricted to observations available before the decision cutoff. All models are evaluated using the same reference answers and scoring rubric.

\noindent\textbf{Evaluation metrics and judge.}
We report answer accuracy (\%) and mean score (0--5), both overall and by interaction category. A locally hosted Qwen3.8-27B serves as a text-only judge, receiving the question, reference answer, model response, and dialogue context when applicable, without access to the source video. The judge first determines binary correctness and then assigns an ordinal score from 0 to 5, with scores of 4--5 indicating correct answers and 0--3 indicating incorrect ones. Contradictory references and unresolved judgments are flagged for review. Judge prompts, parsing rules, and failure handling are detailed in Appendix~\ref{app:judge_prompts}.

\noindent\textbf{Tansfer benchmarks.}
To evaluate cross-benchmark generalization, we compare Qwen2.5-Omni-7B before and after EgoOmniWorld LoRA adaptation on three existing egocentric benchmarks: EgoSound~\cite{zhu2026egosound}, EgoTom~\cite{li2025egotom}, and EgoTempo~\cite{plizzari2025egotempo}. No additional training is performed on these target benchmarks, and accuracy is reported using their respective evaluation protocols.

\subsection{In-domain Evaluation and Training}
\label{sec:in_domain}
We investigate EgoOmniWorld from two perspectives: its diagnostic value
as a unified testbed for identifying capability gaps across Ego--Action--World
interactions, and its effectiveness as a training resource for improving
these capabilities. Table~\ref{tab:main_results} presents overall results
and performance across the four interaction structures, with fine-grained
results for the nine semantic categories provided in
Appendix~\ref{app:category_results}.

\noindent\textbf{Testbed analysis.}
EgoOmniWorld presents substantial challenges for current omni-modal MLLMs. The best closed-sourced model i.e, Gemini 3.8 achieves 61.51\% overall accuracy, while the highest-performing open-source baseline reaches 48.15\%. Among the evaluated 7B-scale models, the best result is only 36.42\%, achieved by Qwen2.5-Omni-7B, highlighting the difficulty of the benchmark.
Beyond overall accuracy, EgoOmniWorld reveals substantial performance variations across interaction structures and semantic categories. In particular, models exhibit different capabilities in recognizing actions, perceiving world states, interpreting intentions, and reasoning about action effects. These variations highlight the diagnostic value of EgoOmniWorld in revealing capability gaps across different components and relationships of the Ego--Action--World interaction process.

\noindent\textbf{Training effectiveness.}
Training on EgoOmniWorld yields substantial overall improvements and broad gains across interaction structures. For Qwen2.5-Omni-7B, LoRA and full fine-tuning (FFT) improve overall accuracy from 36.42\% to 47.36\% and 48.34\%, respectively, with improvements across all four structures and particularly pronounced gains on Edge questions (+13.83 and +15.35 percentage points). Similarly, Qwen3-Omni-Thinking improves from 48.15\% to 52.10\% with LoRA and 53.18\% with FFT, with the latter improving all four structures. The gains vary across models and adaptation strategies: Qwen2.5-Omni-7B benefits more substantially overall, whereas Qwen3-Omni-Thinking shows smaller overall gains, with its LoRA variant slightly reducing Chain and Loop accuracy. These results demonstrate the effectiveness of EgoOmniWorld as interaction-centric supervision while highlighting differences in how model families and training strategies benefit from different interaction structures.

\begin{table}[!htbp]
\centering
\caption{
In-domain performance on EgoOmniWorld.
We report mean score (0--5) and accuracy (\%) across the four interaction structures and overall.
}
\label{tab:main_results}
\begingroup\footnotesize
\setlength{\tabcolsep}{4pt}
\renewcommand{\arraystretch}{1.04}
\begin{tabularx}{\linewidth}{@{}>{\raggedright\arraybackslash}Xccccc@{}}
\toprule
\textbf{Model} & \textbf{Node} & \textbf{Edge} & \textbf{Chain} & \textbf{Loop} & \textbf{Overall} \\
\midrule
\multicolumn{6}{l}{\textcolor{gray}{\textbf{Closed-source models}}} \\
\textcolor{gray}{Gemini 3.8 Flash}
& \textcolor{gray}{2.86/54.74}
& \textcolor{gray}{3.03/60.26}
& \textcolor{gray}{3.34/66.41}
& \textcolor{gray}{3.54/69.29}
& \textcolor{gray}{3.14/61.51} \\
\textcolor{gray}{Qwen3.5-Omni-Plus}
& \textcolor{gray}{2.18/37.31}
& \textcolor{gray}{2.64/50.76}
& \textcolor{gray}{2.80/56.25}
& \textcolor{gray}{2.90/57.89}
& \textcolor{gray}{2.59/49.15} \\
\textcolor{gray}{Qwen3.8-Omni-Flash}
& \textcolor{gray}{2.69/48.62}
& \textcolor{gray}{2.91/56.09}
& \textcolor{gray}{3.06/57.48}
& \textcolor{gray}{3.03/58.27}
& \textcolor{gray}{2.89/54.58} \\
\midrule
\multicolumn{6}{l}{\textbf{Open-source models}} \\
MiniCPM-o 2.6 8B & 2.09/30.28 & 2.01/34.19 & 2.25/38.28 & 2.21/37.83 & 2.11/34.38 \\
SALMONN2+ 72B & 2.16/36.09 & 2.45/44.87 & 2.80/56.25 & 2.67/50.19 & 2.46/44.87 \\
SALMONN2+ 7B & 1.90/30.89 & 2.17/39.49 & 2.02/33.59 & 2.08/35.34 & 2.05/35.28 \\
VideoLLaMA2 7B AV & 1.61/22.94 & 1.56/26.92 & 1.22/16.41 & 1.24/19.48 & 1.46/22.75 \\
Qwen2.5-Omni-3B & 1.97/31.19 & 2.28/42.28 & 2.09/34.38 & 2.07/34.08 & 2.12/36.17 \\
\rowcolor{baserow}Qwen2.5-Omni-7B (base) & 1.90/32.82 & 1.97/37.31 & 2.26/42.37 & 1.93/36.88 & 1.97/36.42 \\
Qwen3-Omni-Instruct & 2.29/40.06 & 2.51/47.09 & 2.59/48.44 & 2.70/52.06 & 2.50/46.37 \\
\rowcolor{baserow}Qwen3-Omni-Thinking (base) & 2.38/42.02 & 2.50/46.68 & 2.80/53.23 & 2.94/55.47 & 2.60/48.15 \\
\midrule
\multicolumn{6}{l}{\textbf{Trained on EgoOmniWorld}} \\
\rowcolor{oursrow}Qwen2.5-Omni-7B-LoRA & 2.58/44.04 & 2.79/51.14 & 2.70/50.78 & 2.48/44.19 & \textbf{2.65/47.36 }\\
\rowcolor{oursrow}Qwen2.5-Omni-7B-FFT & 2.47/41.59 & 2.76/52.66 & 2.69/48.44 & 2.72/50.19 & \textbf{2.66/48.34} \\
\rowcolor{oursrow}Qwen3-Omni-Thinking-LoRA & 2.71/48.93 & 2.83/54.18 & 2.90/52.34 & 2.94/52.81 & \textbf{2.83/52.10} \\
\rowcolor{oursrow}Qwen3-Omni-Thinking-FFT & 2.64/45.57 & 2.90/55.19 & 3.12/60.94 & 2.96/55.81 & \textbf{2.86}/\textbf{53.18} \\
\bottomrule
\end{tabularx}
\endgroup
\end{table}

\subsection{Cross-Benchmark Transfer}
\label{sec:transfer}

\noindent\textbf{Zero-shot transfer.}
We evaluate the transferability of EgoOmniWorld supervision by comparing Qwen2.5-Omni-7B before and after LoRA adaptation on three external egocentric benchmarks: EgoSound, EgoToM, and EgoTempo, without additional training on their respective training sets. As shown in Table~\ref{tab:transfer_results}, adaptation improves accuracy from 31.33\% to 39.41\% on EgoSound, from 55.95\% to 58.22\% on EgoToM, and from 20.40\% to 24.80\% on EgoTempo, corresponding to gains of 8.08, 2.27, and 4.40 percentage points, respectively. These results suggest that the benefits of EgoOmniWorld supervision extend beyond its own benchmark, particularly to audio-visual multimodal egocentric understanding. 

\begin{table}[!htb]
\centering
\caption{
Cross-benchmark transfer accuracy (\%) on existing egocentric benchmarks.
Bold highlights the improved results within our base--LoRA pair.
See Appendix~\ref{app:evaluation_protocol} for evaluation details.
}
\label{tab:transfer_results}

\begingroup
\small
\setlength{\tabcolsep}{4pt}
\renewcommand{\arraystretch}{1.05}
\begin{tabularx}{\linewidth}{@{}>{\raggedright\arraybackslash}Xccc@{}}
\toprule
\textbf{Model} & \textbf{EgoSound} &
\textbf{EgoToM} & \textbf{EgoTempo} \\
\midrule

VideoLLaMA2 7B AV
& 20.5 & -- & -- \\

SALMONN2+ 7B 
& 36.0& -- & 23.60 \\

MiniCPM-o 2.6-8B
& \textbf{40.4} & 57.51 & 20.60\\

\midrule

\rowcolor{baserow}
Qwen2.5-Omni-7B (Base)
& 31.33 & 55.95 & 20.40 \\

\rowcolor{oursrow}
Qwen2.5-Omni-7B + EgoOmniWorld-LoRA
& 39.41 & \textbf{58.22} & \textbf{24.80} \\

Improvement 
& +8.08 & +2.27 & +4.40 \\

\bottomrule
\end{tabularx}
\endgroup
\end{table}

\noindent\textbf{Fine-grained transfer analysis.}
To identify which interaction capabilities benefit from adaptation, we map questions from the three transfer benchmarks to our Ego--Action--World taxonomy (Figure~\ref{fig:transfer_taxonomy}).
On EgoSound, the largest improvement occurs in Ego State, with accuracy increasing from approximately 40\% to 60\%. Substantial gains are also observed in Action Effects, World Perception, and Intent Formation, suggesting benefits across ego-state understanding, perceptual grounding, and action-related reasoning.
On EgoToM, improvements are more modest, but with consistent gains in Ego State, Action Status, and Intent Attribution.
On EgoTempo, improvements are observed in Action Status
and Goal Planning.
Overall, these results show that the benefits of EgoOmniWorld supervision extend across multiple Node and Edge categories.

\noindent\textbf{Fragmented interaction coverage.}
Beyond transfer performance, our taxonomy mapping further reveals the fragmented coverage of existing egocentric benchmarks. Specifically, EgoSound, EgoToM, and EgoTempo each cover only a subset of the Ego--Action--World interaction process, with no explicit evaluation of Chain or Loop in the mapped questions. This highlights the limited coverage of cross-interaction dependencies and feedback-driven adaptation in the benchmarks examined. In contrast, EgoOmniWorld systematically covers all four complementary interaction structures, providing a unified framework for evaluating individual states, directed relations, linked dependencies, and complete Ego--Action--World feedback loops.

\begin{figure}[!htb]
    \centering
    \includegraphics[width=\linewidth]{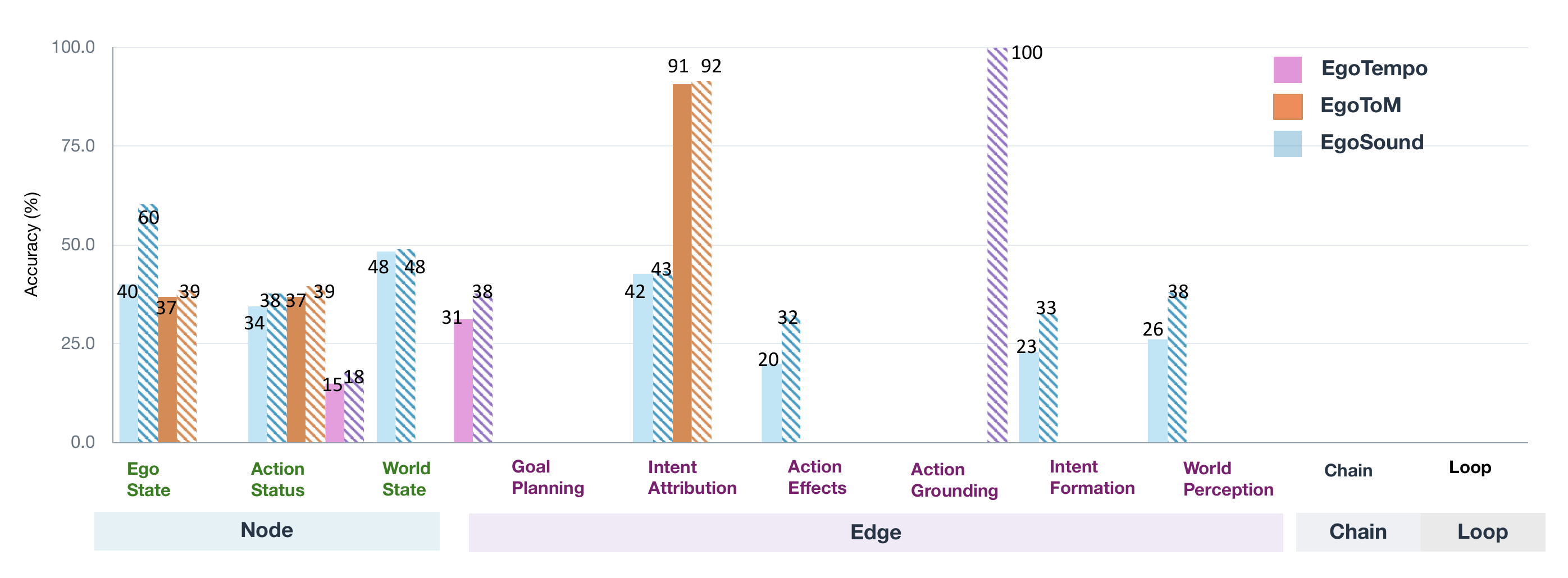}
       \caption{
Fine-grained cross-benchmark transfer analysis under the EgoOmniWorld taxonomy.
We compare Qwen2.5-Omni-7B before (solid bars) and after (hatched bars) EgoOmniWorld LoRA adaptation on EgoSound, EgoToM, and EgoTempo.
Colors distinguish benchmarks, and accuracy is computed on matched questions within each dataset--category group.
Empty categories indicate no evaluated questions, and groups with fewer than three questions are omitted.
}

    \label{fig:transfer_taxonomy}
\end{figure}

\section{Conclusion and Limitations}
We introduced EgoOmniWorld, a unified benchmark that organizes diverse egocentric video understanding capabilities around the closed Ego--Action--World interaction loop. By structuring this process into four complementary scopes---Node, Edge, Chain, and Loop---we provide a common framework for evaluating individual states, directed relations, linked dependencies, and feedback-closed interactions. Our interaction-first, evidence-grounded data engine further integrates visual and auditory evidence and single- and multi-turn QA according to the underlying interaction requirements. Extensive experiments reveal uneven capabilities across existing MLLMs and demonstrate that training on EgoOmniWorld improves in-domain performance and transfers to other egocentric benchmarks. We hope EgoOmniWorld provides a foundation for moving beyond fragmented capability evaluation toward more systematic and holistic understanding of human--world interactions.

\noindent\textbf{Limitations and Future Work.}
While EgoOmniWorld provides a unified framework for egocentric video understanding, several directions remain for future exploration. First, our current dataset is constructed from Ego4D and HD-EPIC. Incorporating more diverse data sources, longer recordings, and richer annotations could further expand its coverage of interaction scenarios and temporal dependencies, potentially improving model generalization and real-world applicability. Second, our training experiments primarily focus on the Qwen-Omni family. Extending EgoOmniWorld supervision to a broader range of omni-modal architectures and training paradigms would further investigate its transferability and benefits for general-purpose egocentric assistants.

\section*{AI Use Statement}
We used large language models (LLMs) throughout the benchmark
construction pipeline to extract structured interactions, ground
multimodal evidence, aggregate related episodes, and generate,
validate, and repair question--answer pairs
(Section~\ref{sec:construction}).
LLMs were also employed as answer judges during evaluation
(Section~\ref{sec:setup}), while a random subset of the test set
underwent manual review (Appendix~\ref{app:human_review}).
AI-assisted tools were additionally used to polish the paper writing. The authors reviewed the AI-assisted outputs and
take full responsibility for the final manuscript, dataset,
and reported findings.

\section*{Ethics Statement}
EgoOmniWorld is constructed from existing Ego4D and HD-EPIC
recordings without collecting new participant videos.
As egocentric recordings may contain identifiable individuals
and private environments, access to and redistribution of
the source videos must comply with the respective datasets'
licenses and access conditions. EgoOmniWorld is intended for research on egocentric interaction understanding. Potential misuse includes unauthorized surveillance or profiling of individuals based on first-person recordings.
Benchmark performance alone should not be interpreted as
evidence of suitability for deployment in privacy-sensitive settings.

\section*{Reproducibility Statement}
Section~\ref{sec:construction} describes the benchmark construction
pipeline, with implementation details and prompt templates provided
in Appendices~\ref{app:construction} and~\ref{app:prompts}.
Section~\ref{sec:setup} details the evaluated models, training
settings, inference protocols, and evaluation metrics.
Additional training configurations, judge prompts, and evaluation
protocols are documented in
Appendices~\ref{app:training_settings},
\ref{app:judge_prompts}, and~\ref{app:evaluation_protocol},
respectively.
We will release the benchmark data, construction and evaluation
code, and trained models.

\FloatBarrier
\bibliographystyle{iclr2027_conference}
\bibliography{references}

@article{cheng2024videgothink,
  title={Videgothink: Assessing egocentric video understanding capabilities for embodied ai},
  author={Cheng, Sijie and Fang, Kechen and Yu, Yangyang and Zhou, Sicheng and Li, Bohao and Tian, Ye and Li, Tingguang and Han, Lei and Liu, Yang},
  journal={arXiv preprint arXiv:2410.11623},
  year={2024}
}

@article{chen2026egoplan,
  title={Egoplan-bench: Benchmarking multimodal large language models for human-level planning},
  author={Chen, Yi and Ge, Yuying and Ge, Yixiao and Ding, Mingyu and Li, Bohao and Wang, Rui and Xu, Ruifeng and Shan, Ying and Liu, Xihui},
  journal={International Journal of Computer Vision},
  volume={134},
  number={3},
  pages={118},
  year={2026},
  publisher={Springer}
}

@article{xu2025tog,
  title={ToG-Bench: Task-Oriented Spatio-Temporal Grounding in Egocentric Videos},
  author={Xu, Qi'ao and Qian, Tianwen and Fu, Yuqian and Li, Kailing and Jiao, Yang and Zhang, Jiacheng and Wang, Xiaoling and He, Liang},
  journal={arXiv preprint arXiv:2512.03666},
  year={2025}
}

@article{mahdi2025exo2egosyn,
  title={Exo2egosyn: Unlocking foundation video generation models for exocentric-to-egocentric video synthesis},
  author={Mahdi, Mohammad and Fu, Yuqian and Savov, Nedko and Pan, Jiancheng and Paudel, Danda Pani and Van Gool, Luc},
  journal={arXiv preprint arXiv:2511.20186},
  year={2025}
}

@article{zhang2023egospot,
  title={EgoSpot: Egocentric Multimodal Control for Hands-Free Mobile Manipulation},
  author={Zhang, Ganlin and Zhang, Deheng and Duan, Longteng and Han, Guo and Fu, Yuqian and Paudel, Danda Pani and Van Gool, Luc and Vollenweider, Eric},
  journal={arXiv preprint arXiv:2306.02393},
  year={2023}
}

@inproceedings{pan2026v,
  title={V\^{} 2-SAM: Marrying SAM2 with Multi-Prompt Experts for Cross-View Object Correspondence},
  author={Pan, Jiancheng and Wang, Runze and Qian, Tianwen and Mahdi, Mohammad and Fu, Yanwei and Xue, Xiangyang and Huang, Xiaomeng and Van Gool, Luc and Paudel, Danda Pani and Fu, Yuqian},
  booktitle={Proceedings of the IEEE/CVF Conference on Computer Vision and Pattern Recognition},
  pages={16910--16919},
  year={2026}
}

@inproceedings{fu2025objectrelator,
  title={Objectrelator: Enabling cross-view object relation understanding across ego-centric and exo-centric perspectives},
  author={Fu, Yuqian and Wang, Runze and Ren, Bin and Sun, Guolei and Gong, Biao and Fu, Yanwei and Paudel, Danda Pani and Huang, Xuanjing and Van Gool, Luc},
  booktitle={2025 IEEE/CVF International Conference on Computer Vision (ICCV)},
  pages={6530--6540},
  year={2025},
  organization={IEEE}
}

@inproceedings{ye2025mmego,
  title={MMEgo: Towards building egocentric multimodal LLMs for video QA},
  author={Ye, Hanrong and Zhang, Haotian and Daxberger, Erik and Chen, Lin and Lin, Zongyu and Li, Yanghao and Zhang, Bowen and You, Haoxuan and Xu, Dan and Gan, Zhe and others},
  booktitle={International Conference on Learning Representations},
  volume={2025},
  pages={71705--71723},
  year={2025}
}

@article{pan2026egointent,
  title={EgoIntent: An Egocentric Step-level Benchmark for Understanding What, Why, and Next},
  author={Pan, Ye and Wong, Chi Kit and Lyu, Yuanhuiyi and Li, Hanqian and Huo, Jiahao and Chen, Jiacheng and Jiang, Lutao and Zheng, Xu and Hu, Xuming},
  journal={arXiv preprint arXiv:2603.12147},
  year={2026}
}

@inproceedings{rodin2025easg,
  title={Easg-bench: Video q\&a benchmark with egocentric action scene graphs},
  author={Rodin, Ivan and Wu, Tz-Ying and Min, Kyle and Sridhar, Sharath Nittur and Furnari, Antonino and Tripathi, Subarna and Farinella, Giovanni Maria},
  booktitle={2025 IEEE/CVF International Conference on Computer Vision Workshops (ICCVW)},
  pages={2732--2737},
  year={2025},
  organization={IEEE}
}

@inproceedings{dong2026visnec,
  title={Visnec: Measuring and leveraging visual necessity for multimodal instruction tuning},
  author={Dong, Mingkang and Cai, Hongyi and Li, Jie and Zhou, Sifan and Ren, Bin and Peng, Kunyu and Fu, Yuqian},
  booktitle={European Conference on Computer Vision},
  pages={531--548},
  year={2026},
  organization={Springer}
}

@inproceedings{grauman2022ego4d,
  title={Ego4d: Around the world in 3,000 hours of egocentric video},
  author={Grauman, Kristen and Westbury, Andrew and Byrne, Eugene and Chavis, Zachary and Furnari, Antonino and Girdhar, Rohit and Hamburger, Jackson and Jiang, Hao and Liu, Miao and Liu, Xingyu and others},
  booktitle={Proceedings of the IEEE/CVF conference on computer vision and pattern recognition},
  pages={18995--19012},
  year={2022}
}

@inproceedings{perrett2025hd,
  title={Hd-epic: A highly-detailed egocentric video dataset},
  author={Perrett, Toby and Darkhalil, Ahmad and Sinha, Saptarshi and Emara, Omar and Pollard, Sam and Parida, Kranti Kumar and Liu, Kaiting and Gatti, Prajwal and Bansal, Siddhant and Flanagan, Kevin and others},
  booktitle={2025 IEEE/CVF Conference on Computer Vision and Pattern Recognition (CVPR)},
  pages={23901--23913},
  year={2025},
  organization={IEEE}
}

@article{Xu2025Qwen25OmniTR,
  title={Qwen2.5-Omni Technical Report},
  author={Jin Xu and Zhifang Guo and Jinzheng He and Hang-Rui Hu and Ting He and Shuai Bai and Ke-qin Chen and Jia-Lin Wang and Yang Fan and Kai Dang and Bin Zhang and Xiong Wang and Yunfei Chu and Junyang Lin},
  journal={ArXiv},
  year={2025},
  volume={abs/2503.20215},
  url={https://api.semanticscholar.org/CorpusID:277322543}
}

@article{gao2026lifeeval,
  title={LifeEval: A Multimodal Benchmark for Assistive AI in Egocentric Daily Life Tasks},
  author={Gao, Hengjian and Zhang, Kaiwei and Wang, Shibo and Chen, Mingjie and Cao, Qihang and Wang, Xianfeng and Zhu, Yucheng and Min, Xiongkuo and Sun, Wei and Zhu, Dandan and others},
  journal={arXiv preprint arXiv:2603.00490},
  year={2026}
}

@inproceedings{zhu2026egosound,
  title={Egosound: Benchmarking sound understanding in egocentric videos},
  author={Zhu, Bingwen and Fu, Yuqian and Dong, Qiaole and Sun, Guolei and Qian, Tianwen and Wu, Yuzheng and Paudel, Danda Pani and Fu, Yanwei and Xue, Xiangyang},
  booktitle={Proceedings of the IEEE/CVF Conference on Computer Vision and Pattern Recognition},
  pages={25589--25598},
  year={2026}
}

@article{li2025egotom,
  title={Egotom: Benchmarking theory of mind reasoning from egocentric videos},
  author={Li, Yuxuan and Veerabadran, Vijay and Iuzzolino, Michael L and Roads, Brett D and Celikyilmaz, Asli and Ridgeway, Karl},
  journal={arXiv preprint arXiv:2503.22152},
  year={2025}
}

@article{jia2022egotaskqa,
  title={Egotaskqa: Understanding human tasks in egocentric videos},
  author={Jia, Baoxiong and Lei, Ting and Zhu, Song-Chun and Huang, Siyuan},
  journal={Advances in Neural Information Processing Systems},
  volume={35},
  pages={3343--3360},
  year={2022}
}

@article{cheng2024videollama,
  title={Videollama 2: Advancing spatial-temporal modeling and audio understanding in video-llms},
  author={Cheng, Zesen and Leng, Sicong and Zhang, Hang and Xin, Yifei and Li, Xin and Chen, Guanzheng and Zhu, Yongxin and Zhang, Wenqi and Luo, Ziyang and Zhao, Deli and others},
  journal={arXiv preprint arXiv:2406.07476},
  year={2024}
}

@article{sun2025video,
  title={video-salmonn-o1: Reasoning-enhanced audio-visual large language model},
  author={Sun, Guangzhi and Yang, Yudong and Zhuang, Jimin and Tang, Changli and Li, Yixuan and Li, Wei and Ma, Zejun and Zhang, Chao},
  journal={arXiv preprint arXiv:2502.11775},
  year={2025}
}

@article{cui2026minicpm,
  title={Minicpm-o 4.5: Towards real-time full-duplex omni-modal interaction},
  author={Cui, Junbo and Xu, Bokai and Wang, Chongyi and Yu, Tianyu and Sun, Weiyue and Xu, Yingjing and Wang, Tianran and He, Zhihui and Ma, Wenshuo and Cai, Tianchi and others},
  journal={arXiv preprint arXiv:2604.27393},
  year={2026}
}

@inproceedings{zhou2025egotextvqa,
  title={Egotextvqa: Towards egocentric scene-text aware video question answering},
  author={Zhou, Sheng and Xiao, Junbin and Li, Qingyun and Li, Yicong and Yang, Xun and Guo, Dan and Wang, Meng and Chua, Tat-Seng and Yao, Angela},
  booktitle={2025 IEEE/CVF Conference on Computer Vision and Pattern Recognition (CVPR)},
  pages={3363--3373},
  year={2025},
  organization={IEEE}
}

@inproceedings{zhang2026egonight,
  title={Egonight: Towards egocentric vision understanding at night with a challenging benchmark},
  author={Zhang, Deheng and Fu, Yuqian and Yang, Runyi and Miao, Yang and Qian, Tianwen and Zheng, Xu and Sun, Guolei and Chhatkuli, Ajad and Huang, Xuanjing and Jiang, Yu-Gang and others},
  booktitle={International Conference on Learning Representations},
  volume={2026},
  pages={887--901},
  year={2026}
}

@inproceedings{li2026egocross,
  title={Egocross: Benchmarking multimodal large language models for cross-domain egocentric video question answering},
  author={Li, Yanjun and Fu, Yuqian and Qian, Tianwen and Xu, Qi'ao and Dai, Silong and Paudel, Danda Pani and Van Gool, Luc and Wang, Xiaoling},
  booktitle={Proceedings of the AAAI Conference on Artificial Intelligence},
  volume={40},
  number={8},
  pages={6592--6600},
  year={2026}
}

@inproceedings{chen2025grounded,
  title={Grounded multi-hop videoqa in long-form egocentric videos},
  author={Chen, Qirui and Di, Shangzhe and Xie, Weidi},
  booktitle={Proceedings of the AAAI Conference on Artificial Intelligence},
  volume={39},
  number={2},
  pages={2159--2167},
  year={2025}
}

@inproceedings{yang2025egolife,
  title={Egolife: Towards egocentric life assistant},
  author={Yang, Jingkang and Liu, Shuai and Guo, Hongming and Dong, Yuhao and Zhang, Xiamengwei and Zhang, Sicheng and Wang, Pengyun and Zhou, Zitang and Xie, Binzhu and Wang, Ziyue and others},
  booktitle={2025 IEEE/CVF Conference on Computer Vision and Pattern Recognition (CVPR)},
  pages={28885--28900},
  year={2025},
  organization={IEEE}
}

@article{team2026qwen3,
  title={Qwen3. 5-omni technical report},
  author={Team, Qwen},
  journal={arXiv preprint arXiv:2604.15804},
  year={2026}
}

@article{team2023gemini,
  title={Gemini: a family of highly capable multimodal models},
  author={Team, Gemini and Anil, Rohan and Borgeaud, Sebastian and Alayrac, Jean-Baptiste and Yu, Jiahui and Soricut, Radu and Schalkwyk, Johan and Dai, Andrew M and Hauth, Anja and Millican, Katie and others},
  journal={arXiv preprint arXiv:2312.11805},
  year={2023}
}

@article{mangalam2023egoschema,
  title={Egoschema: A diagnostic benchmark for very long-form video language understanding},
  author={Mangalam, Karttikeya and Akshulakov, Raiymbek and Malik, Jitendra},
  journal={Advances in Neural Information Processing Systems},
  volume={36},
  pages={46212--46244},
  year={2023}
}

@inproceedings{plizzari2025omnia,
  title={Omnia de egotempo: Benchmarking temporal understanding of multi-modal llms in egocentric videos},
  author={Plizzari, Chiara and Tonioni, Alessio and Xian, Yongqin and Kulshrestha, Achin and Tombari, Federico},
  booktitle={2025 IEEE/CVF Conference on Computer Vision and Pattern Recognition (CVPR)},
  pages={24129--24138},
  year={2025},
  organization={IEEE}
}

@inproceedings{plizzari2025egotempo,
  title = {Omnia de {EgoTempo}: Benchmarking Temporal Understanding of Multi-Modal {LLMs} in Egocentric Videos},
  author = {Plizzari, Chiara and Tonioni, Alessio and Xian, Yongqin and Kulshrestha, Achin and Tombari, Federico},
  booktitle = {Proceedings of the IEEE/CVF Conference on Computer Vision and Pattern Recognition},
  year = {2025},
  url = {https://arxiv.org/abs/2503.13646}
}

@misc{seth2026egoavu,
  title = {{EgoAVU}: Egocentric Audio-Visual Understanding},
  author = {Seth, Ashish and Mei, Xinhao and Zhao, Changsheng and Nagaraja, Varun and Chang, Ernie and Meyer, Gregory P. and Le Lan, Gael and Xiong, Yunyang and Chandra, Vikas and Shi, Yangyang and Manocha, Dinesh and Cai, Zhipeng},
  year = {2026},
  eprint = {2602.06139},
  archivePrefix = {arXiv},
  primaryClass = {cs.CV},
  url = {https://arxiv.org/abs/2602.06139}
}
\clearpage
\appendix

\section{Dataset Statistics and Representative Examples}
\label{app:dataset_statistics}
This section reports dataset composition, temporal coverage, and dialogue statistics, followed by representative QA examples. A single-turn QA or a complete multi-turn dialogue counts as one QA group throughout.

\noindent\textbf{Scale and task distribution.}
EgoOmniWorld contains 24,166 QA groups, comprising 20,284 training and 3,882 test samples. A QA group is either a standalone question or a complete multi-turn dialogue. Node, Edge, Chain, and Loop contain 10,026, 9,521, 2,480, and 2,139 samples, respectively, representing 41.49\%, 39.40\%, 10.26\%, and 8.85\% of the total. Node covers ego states, action dynamics, and world states, while Edge covers six directed interaction relations. Chain and Loop assess open relational paths and interactions with feedback closure, respectively (Table~\ref{tab:dataset_counts}; Figure~\ref{fig:dataset_sta}).

\begin{table}[htbp]
\caption{Sample counts by task category and split. A multi-turn dialogue counts as one QA group.}
\label{tab:dataset_counts}
\centering
\begin{tabular}{llrrr}
\toprule
Level & Category & Train & Test & Total \\
\midrule
Node & Ego State & 1,911 & 345 & 2,256 \\
 & Action Status & 4,549 & 890 & 5,439 \\
 & World State & 1,966 & 365 & 2,331 \\
 & Subtotal & 8,426 & 1,600 & 10,026 \\
\midrule
Edge & Goal Planning & 242 & 40 & 282 \\
 & Intent Attribution & 644 & 127 & 771 \\
 & Action Effects & 1,246 & 235 & 1,481 \\
 & Action Grounding & 1,817 & 334 & 2,151 \\
 & Intent Formation & 681 & 98 & 779 \\
 & World Perception & 3,387 & 670 & 4,057 \\
 & Subtotal & 8,017 & 1,504 & 9,521 \\
\midrule
Chain & & 2,171 & 309 & 2,480 \\
Loop & & 1,670 & 469 & 2,139 \\
\midrule
\textbf{Total} & & \textbf{20,284} & \textbf{3,882} & \textbf{24,166} \\
\bottomrule
\end{tabular}
\end{table}

\noindent\textbf{Video duration.}
Duration statistics cover the 23,246 samples with clip metadata, excluding the 920 restored samples for which new clips have not been generated. Deduplication by source video ID and temporal interval yields 10,664 clips: 8,892 in training and 1,772 in testing. According to the clip metadata, the mean and median durations are 57.2 and 34.5 seconds; 92.7\% of clips are no longer than two minutes, and the longest lasts approximately 46.7 minutes. When weighted by QA groups within this subset, retaining repeated references to the same clip, the mean duration is 71.5 seconds and the median is 54.1 seconds. Samples associated with 30--60-second and 60--120-second clips account for 44.7\% and 28.8\%, respectively. The data therefore primarily cover short interactions while retaining samples that require understanding over longer temporal spans (Figure~\ref{fig:dataset_sta}).

\noindent\textbf{Question formats and dialogue settings.}
The dataset supports multiple-choice and open-ended questions. It contains 19,833 single-turn groups and 4,333 multi-turn dialogues, representing 82.07\% and 17.93\%, respectively. All multi-turn dialogues use open-ended answers.

\noindent\textbf{Multi-turn dialogue length.}
The 4,333 multi-turn dialogues comprise 11,273 QA turns, averaging 2.60 turns per dialogue. Two-turn dialogues account for 1,750 samples (40.39\%), three-turn dialogues for 2,559 (59.06\%), and four-turn dialogues for 24 (0.55\%). The data are dominated by compact two- and three-turn dialogues, which use successive questions to examine understanding of related facts and reasoning dependencies (Figure~\ref{fig:dataset_sta}).

\subsection{Representative QA Examples}
\label{app:qa_examples}
Figure~\ref{fig:appendix_22examples} presents 22 examples: two for each of the nine Node and Edge semantic categories, two Chain dialogues, and two Loop dialogues. The examples pair questions and reference answers with frames from the corresponding clips. Chain examples expose dependencies between earlier and later events; Loop examples show how an observed outcome informs a subsequent action. Audio-visual examples additionally refer to sounds that cannot be conveyed by still frames alone.

\clearpage
\begin{figure}[p]
  \centering
  \includegraphics[width=\linewidth,height=0.85\textheight,keepaspectratio]{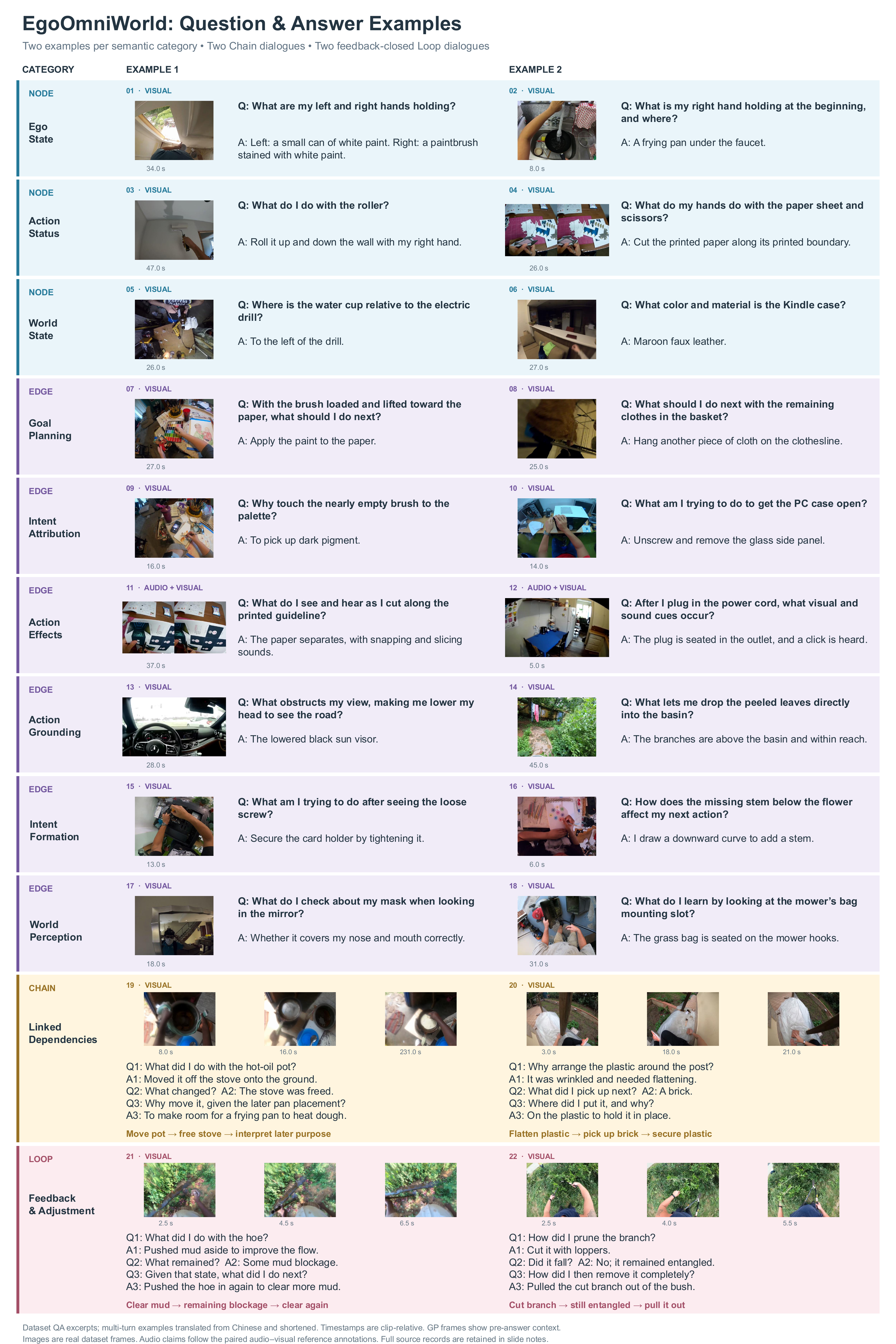}
  \caption{Representative QA examples in EgoOmniWorld. The figure contains two examples for each of Ego State, Action Status, World State, Goal Planning, Intent Attribution, Action Effects, Action Grounding, Intent Formation, and World Perception, together with two Chain and two Loop dialogues. Frames come from the corresponding dataset clips; timestamps are clip-relative. Displayed questions and answers are shortened for readability, and Chinese multi-turn examples are translated into English. Each complete dialogue counts as one QA group. The examples illustrate the annotations rather than establish exhaustive validation of their assigned categories.}
  \label{fig:appendix_22examples}
\end{figure}
\clearpage

\section{Data Construction Details}
\label{app:construction}
This appendix provides implementation details omitted from the main text, including video selection, interaction representations, category-specific QA generation, structural validation, constrained repair, and evidence provenance. The organization follows the interaction-first pipeline in Section~\ref{sec:construction}.

\subsection{Source Videos and Provenance}
Ego4D source selection uses narration-based clustering to promote activity diversity; the implementation and observed coverage appear below. HD-EPIC segments use official narrations and action intervals for candidate localization. Narrations are temporal anchors and provenance references, not substitutes for supporting evidence from the original audio and video.

\subsection{Interaction representation details}
\noindent\textbf{Local interaction components.}
For each candidate interaction, the engine extracts a structured record that includes the following components when supported by evidence:
\begin{itemize}
\item the pre-action world state;
\item observations that trigger or contextualize the interaction;
\item the current intention or persistent activity goal;
\item the current decision or intermediate subgoal;
\item the action actually performed;
\item the resulting change in world state;
\item perceptual or task-level feedback; and
\item attention-related information when sufficiently supported.
\end{itemize}

We distinguish these components by their functional roles. An \emph{intention} is a relatively persistent goal, a \emph{subgoal} is the currently selected intermediate step, and an \emph{action} is an observable behavior that has been performed. In particular, a paraphrase of what the wearer is currently doing is not automatically an intention description.

For latent components such as intentions and decisions, records also preserve supporting observations, speech, or contextual information. Unsupported latent states remain unspecified rather than being completed through commonsense inference.

\noindent\textbf{Multimodal perceptual enrichment.}
Visual evidence may describe object identities and attributes, spatial layouts, hand occupancy, hand--object interactions, the wearer's surroundings, visible state changes, and changes in location. Auditory evidence may include speech, environmental sounds, action-generated sounds, sound-source identities, and relative source directions when available.

Each perceptual record is associated with its temporal interval and modality. When visual and auditory observations are considered jointly, the pipeline also records their correspondence within the interaction instead of assuming that temporal proximity implies semantic relevance.

\noindent\textbf{Component relations.}
The resulting local representation is a component graph whose nodes represent interaction components and whose edges represent candidate relations. Connectivity is treated as a structural hypothesis, not proof of causality. Relations used for QA generation must additionally pass the evidence checks for the relevant category.

\subsection{Entity Association and Episode Dependencies}
\noindent\textbf{Object identity association.}
Object observations from different local interactions are associated only when they can reliably be identified as referring to the same physical entity. The resulting track stores observations in temporal order, including their states, locations, associated interactions, timestamps, and evidence sources.

Only entity identity is merged; states observed at different times remain separate records. A track therefore preserves successive changes in an object's state without collapsing temporally distinct observations into a single description.

When identity cannot be reliably established, records remain separate. Similar names, shared semantic categories, or temporal proximity are insufficient grounds for merging.

\noindent\textbf{Macro-level task aggregation.}
Local interactions are processed in temporally overlapping windows. Adjacent interactions that serve the same identifiable activity stage are grouped into a macro-level task episode. A new episode is created when the goal, principal object set, workspace, or activity changes substantially.

Each macro episode preserves its source events and, where recorded, its persistent intention, current subgoals, initial and resulting states, key entities, instructions, constraints, failures, interruptions, and unfinished tasks.

Episodes produced by overlapping windows are then reconciled. Duplicate episode descriptions are merged while preserving temporal order and event provenance.

\noindent\textbf{Cross-episode relations.}
Candidate relations between macro episodes include state transfer, instruction following, failure correction, resumption of unfinished tasks, environmental constraints, and continuation of persistent goals as subgoals change.

Each relation stores its source and target episodes, relevant source events, connected interaction components, relation type, transferred information or state, required modalities, and uncertainty metadata. Relations with unresolved entity identity, state transfer, or semantic dependence are not used directly to generate questions.

\subsection{Ego4D Clustering and Source Selection}
\label{app:clustering}
The statistics below concern source-video selection, not QA counts. They match saved clustering outputs to the final selected video identifiers without rerunning clustering.

\noindent\textbf{Video-level narration representation and clustering.}
The clustering implementation combines summaries and action descriptions from the two available rounds of narrations for each video into a video-level text representation. Text cleaning removes markup, normalizes whitespace, and deduplicates repeated descriptions before concatenating summaries and action descriptions. TF--IDF features, truncated SVD, L2 normalization, and MiniBatch K-Means then group videos with similar narration content.

The original clustering summary records the method as \texttt{tfidf\_svd\_minibatch\_kmeans}, with random seed 42, covering 8,931 videos and 160 clusters. Each video retains its original identifier, cluster index, and distance to the corresponding cluster center. Cluster summaries also store frequent keywords and representative videos near the center to facilitate inspection and topic interpretation. Keywords aid interpretation; they are neither human activity labels nor validated semantic categories.

\noindent\textbf{Candidate filtering and within-cluster ranking.}
The sampler first selects candidates with complete annotation status, valid video identifiers, and durations within the specified range. Importantly, \texttt{duration\_proxy\_sec} uses the maximum timestamp across the two narration rounds as a duration proxy rather than directly reading the media file's duration. The filter therefore measures the temporal span covered by narrations.

The sampling budget first provides minimum coverage quotas for available clusters, then allocates additional places according to remaining candidate capacity, never exceeding a cluster's capacity. Candidates are ranked primarily by ascending distance to the center, with descending narration density as a tie-breaker and video identifiers providing deterministic ordering. Narration density is computed as the narration count divided by the duration proxy floored at 60 seconds.

The saved sampling script defaults to a total budget of 1,000, a minimum quota of four per cluster, and a duration-proxy range of 180--3,600 seconds. These are verifiable script defaults, not confirmed historical execution parameters: the complete parameter record for the original run was not found during this audit. The observed distribution below comes directly from the final selected list and does not depend on assumptions about run parameters.

\noindent\textbf{Observed selection results and coverage.}
\begin{table}[htbp]
\caption{Observed Ego4D source-video selection statistics.}
\label{tab:selection_stats}
\centering
\begin{tabular}{lr}
\toprule Statistic & Value \\ \midrule
Videos in the original clustering & 8,931 \\
Original clusters & 160 \\
Selected source videos (deduplicated) & 1,000 \\
Selected videos matched to original clusters & 1,000 \\
Clusters covered & 159 \\
Cluster coverage & 99.375\% \\
Minimum/maximum selections per covered cluster & 2/12 \\
Clusters with five or six selected videos & 102 \\
\bottomrule
\end{tabular}
\end{table}

Cluster 39 is the only cluster with no selected video. It originally contains a single video whose narration-based duration proxy is approximately 92.3 seconds. This is consistent with the current script's default exclusion of candidates below 180 seconds, but the historical exclusion reason cannot be established from default parameters alone. Clusters 7 and 112 have the most selected videos, with 12 each.

Cluster-stratified sampling aims to reduce budget domination by frequent activities under sampling based only on video counts, balancing activity coverage and within-cluster representativeness. It is not an unbiased estimate of the natural activity distribution and does not guarantee coverage of every fine-grained activity. Center-based ranking may reduce the inclusion probability of peripheral or rare activities; variations in narration wording may also distribute similar activities across multiple clusters. Coverage of 159/160 therefore refers only to the original clustering and does not mean that 159 mutually exclusive human-defined activity categories are covered.

\noindent\textbf{Cluster topics and selected-video distribution.}
The following table preserves the original cluster indices and counts. Topic labels are translated from the Chinese summaries derived from original keywords and representative narrations; they are provided for readability, not as original human annotations. Heterogeneous clusters are marked ``mixed.'' For example, crochet, yarn crochet, and hook-based crochet may occur in different clusters without constituting mutually exclusive activity categories. ``All'' denotes the original cluster size, while ``Selected'' denotes the number of videos from the final 1,000-video subset in that cluster. The first six keywords are retained from the original summaries.

\begingroup
\small
\setlength{\tabcolsep}{3pt}
\begin{longtable}{@{}r p{0.27\linewidth} r r p{0.36\linewidth}@{}}
\caption{Original Ego4D clusters and selected source videos. Topic summaries are interpretive, not human-annotated activity classes.}\label{tab:clusters}\\
\toprule
ID & Topic & All & Selected & Original keywords (first six)\\
\midrule
\endfirsthead
\toprule
ID & Topic & All & Selected & Original keywords (first six)\\
\midrule
\endhead
\bottomrule
\endfoot
0 & Crochet & 10 & 5 & yarn, fabric, hand, crochet, left, right \\
1 & Kitchen weighing and container washing & 11 & 4 & hand, table, sink, jug, summary, kitchen \\
2 & Paper crafts and gluing & 44 & 6 & paper, craft, table, hand, right, drops \\
3 & Multiplayer card games & 148 & 10 & man, table, hand, cards, right, card \\
4 & Reading and bookshelf organization & 45 & 6 & book, books, shelf, summary, floor, phone \\
5 & Supermarket shopping & 95 & 10 & shopping, basket, man, woman, bag, paper \\
6 & Knitting & 58 & 5 & hand, yarn, knitting, right, fabric, knitted \\
7 & Indoor manual activities and chores (mixed) & 214 & 12 & hand, right, left, hands, bamboo, summary \\
8 & Painting and coloring on a board & 75 & 7 & paint, brush, hand, painting, board, right \\
9 & Metal-part assembly and fastening & 37 & 5 & box, unsure, screw, metal, table, hand \\
10 & Writing notes & 44 & 5 & book, hand, table, right, pen, left \\
11 & Clearing branches from fences & 24 & 5 & hand, left, right, twig, wire, pulls \\
12 & Woodwork and floor installation & 86 & 7 & wood, hand, floor, tape, tile, right \\
13 & Bicycle tire repair and inflation & 9 & 5 & bicycle, tire, pump, tyre, hand, wheel \\
14 & Pencil drawing and sketching & 62 & 7 & hand, pencil, paper, table, book, right \\
15 & Gardening and pruning & 79 & 6 & hand, plant, right, ground, left, flower \\
16 & Hand sewing and embroidery & 52 & 6 & thread, hand, needle, cloth, right, summary \\
17 & Browsing and talking in shops & 12 & 5 & woman, man, magazine, summary, phone, unsure \\
18 & Dog walking & 67 & 7 & dog, hand, leash, road, left, right \\
19 & Cooking and ingredient preparation & 55 & 6 & hand, right, left, kitchen, spoon, drops \\
20 & Walking on roads & 42 & 5 & road, walked, summary, hand, man, woman \\
21 & Drum-kit playing & 69 & 6 & drum, hand, right, left, set, hands \\
22 & Motorcycle and mechanical-part repair & 111 & 10 & hand, right, left, screw, table, hands \\
23 & Outdoor climbing and rope handling & 48 & 5 & man, rope, mountain, hand, rock, hands \\
24 & Card and board games & 131 & 9 & cards, man, woman, table, card, summary \\
25 & Washing and hanging laundry & 62 & 6 & machine, washing, cloth, clothes, laundry, hand \\
26 & Driving & 70 & 8 & car, hand, drives, right, left, past \\
27 & Lawnmower engine repair & 109 & 8 & hand, mower, lawn, right, left, unsure \\
28 & Grass cutting with handheld trimmers & 55 & 6 & grass, hand, cutter, trimmer, left, right \\
29 & Reading books & 76 & 8 & book, hand, summary, page, read, right \\
30 & Dishwashing and utensil cleaning & 41 & 5 & sink, plate, hand, sponge, kitchen, drops \\
31 & Glue-gun and cord crafts & 39 & 5 & hand, right, left, glue, floor, gun \\
32 & Shop browsing and conversation & 95 & 10 & man, woman, phone, store, summary, shelf \\
33 & Kitchen retrieval and daily activities (mixed) & 110 & 9 & unsure, kitchen, opens, fridge, summary, cabinet \\
34 & Stirring food and mixing dough & 56 & 5 & hand, right, left, bowl, pot, drops \\
35 & Peeling and cutting produce & 52 & 6 & hand, right, left, tray, knife, drops \\
36 & Dominoes and indoor activities (mixed) & 55 & 6 & man, table, hand, game, summary, board \\
37 & Wall painting with rollers and brushes & 106 & 10 & paint, hand, brush, roller, right, wall \\
38 & Flatbread preparation and pan cooking & 45 & 5 & dough, pan, hand, spoon, oil, right \\
39 & Carpet placement and adjustment & 1 & 0 & carpet, man, pulls, floor, helps, turns \\
40 & Hook-based crochet & 68 & 7 & crochet, yarn, hand, hook, fabric, left \\
41 & Ironing and folding clothes & 36 & 5 & table, cloth, iron, hand, hands, right \\
42 & Street walking and running & 54 & 6 & road, summary, walked, phone, hand, sidewalk \\
43 & Group hiking and outdoor activities & 51 & 4 & woman, hand, man, summary, right, left \\
44 & Shared cooking and kitchen conversation & 118 & 10 & hand, right, man, left, kitchen, unsure \\
45 & Cleaning and storing kitchen utensils & 81 & 8 & sink, kitchen, washes, places, summary, bowl \\
46 & Wooden mechanical-model assembly & 74 & 8 & table, piece, hand, wood, wooden, model \\
47 & Nailing and hammering in woodwork & 41 & 5 & hand, hammer, right, wood, nail, left \\
48 & Reading music and piano practice & 30 & 5 & hand, piano, left, right, book, hands \\
49 & Metal welding and grinding & 73 & 6 & hand, metal, right, left, man, summary \\
50 & Electronic components and cable handling & 22 & 5 & hand, cable, right, table, left, unsure \\
51 & Chainsaw maintenance & 8 & 5 & hand, right, chainsaw, left, saw, unsure \\
52 & Hand sanding wood & 49 & 5 & hand, sandpaper, left, right, wood, paper \\
53 & Branch pruning & 62 & 7 & hand, tree, right, left, branch, hands \\
54 & Pottery and clay shaping & 74 & 6 & hand, clay, right, mold, brick, left \\
55 & Kitchen preparation and retrieval (mixed) & 40 & 6 & hand, kitchen, drops, sink, right, cabinet \\
56 & Tablet and electronic-device use & 29 & 5 & tablet, hand, right, phone, summary, screen \\
57 & Lawn mowing and grass-clipping removal & 46 & 6 & mower, lawn, grass, bag, hand, summary \\
58 & Watching television & 75 & 7 & hand, television, man, summary, watched, room \\
59 & Cleaning and organizing bookshelves & 45 & 5 & book, hand, books, right, left, shelf \\
60 & Guitar playing & 76 & 6 & guitar, hand, played, right, summary, room \\
61 & Painting and color mixing & 105 & 10 & paint, brush, table, hand, paper, painting \\
62 & Kitchen preparation and family interaction & 75 & 8 & hand, woman, right, left, unsure, table \\
63 & Wood spray-painting and spray-gun use & 18 & 5 & paint, hand, sprayer, gun, spray, wood \\
64 & Hoeing and field clearing & 51 & 5 & hand, hoe, ground, man, right, hands \\
65 & Car-engine repair and oil changes & 59 & 6 & car, hand, unsure, engine, right, box \\
66 & Ironing shirts and clothes & 20 & 5 & shirt, iron, table, hand, cloth, board \\
67 & Dishwashing and kitchen organization & 76 & 8 & hand, right, left, sink, kitchen, drops \\
68 & Tracing and cutting paper & 64 & 6 & paper, hand, table, right, left, piece \\
69 & Wall and ceiling renovation & 48 & 5 & hand, wall, paint, ladder, scraper, wallpaper \\
70 & Walking and talking with others & 36 & 5 & summary, man, woman, walked, hand, looked \\
71 & Kitchen chopping and meal preparation & 103 & 10 & drops, board, kitchen, unsure, hand, sink \\
72 & Watering and tending potted plants & 22 & 5 & plant, flower, hand, water, pot, bottle \\
73 & Drilling and assembly & 66 & 6 & hand, drill, right, table, left, wall \\
74 & Bicycle-component repair & 76 & 7 & hand, right, bicycle, left, table, unsure \\
75 & Drawing and coloring with pens & 70 & 8 & pen, hand, drawing, left, paper, book \\
76 & Chess & 17 & 5 & chess, man, piece, hand, board, summary \\
77 & Yarn crochet & 14 & 5 & hand, crochet, right, left, wool, hook \\
78 & Bag handling and camping storage (mixed) & 38 & 5 & bag, man, hand, right, table, tent \\
79 & Room tidying and item storage & 100 & 9 & hand, right, left, unsure, hands, floor \\
80 & Parking-lot walking and socializing (mixed) & 13 & 5 & man, parking, lot, road, summary, walked \\
81 & Clothes shopping and browsing & 72 & 9 & clothes, man, cloth, hanger, woman, rack \\
82 & Lawn mowing with push mowers & 21 & 5 & lawn, mower, hand, right, grass, left \\
83 & Desktop computer and keyboard/mouse use & 53 & 5 & hand, computer, mouse, summary, phone, keyboard \\
84 & Reading and indoor activities & 24 & 5 & book, books, summary, shelf, wall, room \\
85 & Ball sports (mixed) & 56 & 6 & ball, man, tennis, summary, throws, field \\
86 & Piping and decoration & 25 & 5 & hand, right, bag, bowl, left, dough \\
87 & Wall measurement and paper positioning & 33 & 5 & wall, tape, paper, piano, measure, summary \\
88 & Field activities and conversation (mixed) & 23 & 3 & man, hand, grass, unsure, room, summary \\
89 & Hedge and shrub trimming & 56 & 6 & hand, trimmer, hedge, right, hands, left \\
90 & Interacting with cats & 22 & 5 & cat, hand, right, floor, summary, room \\
91 & Laptop and phone use & 81 & 7 & laptop, hand, phone, table, summary, right \\
92 & Laboratory tubes and sample handling & 29 & 5 & hand, summary, right, tube, test, table \\
93 & Washing and preparing ingredients & 49 & 6 & sink, bowl, drops, unsure, spoon, board \\
94 & Yarn crochet & 55 & 6 & hand, yarn, crochet, right, left, hook \\
95 & Wood sawing and planing & 124 & 10 & hand, wood, right, table, left, saw \\
96 & Indoor conversation and socializing (mixed) & 71 & 7 & man, hand, table, woman, hands, right \\
97 & Bicycle riding & 127 & 10 & bicycle, road, summary, man, rides, hand \\
98 & Vacuum cleaning & 70 & 6 & vacuum, cleaner, hand, floor, summary, left \\
99 & Hose watering and spraying & 47 & 5 & hand, right, left, hose, pipe, water \\
100 & Fabric weaving and crochet & 81 & 8 & thread, fabric, crochet, hand, summary, knitted \\
101 & Construction and cement work & 40 & 5 & man, cement, hand, wall, bucket, trowel \\
102 & Cooking and stirring in pots & 65 & 6 & spoon, pot, hand, cooking, unsure, kitchen \\
103 & Small-engine and lawnmower maintenance & 17 & 5 & hand, right, left, oil, mower, lawn \\
104 & Brick and concrete-block laying & 41 & 5 & brick, bricks, hand, ground, summary, concrete \\
105 & Car tire and wheel repair & 98 & 9 & car, wheel, unsure, tire, hand, garage \\
106 & Playing fetch and interacting with dogs & 38 & 5 & dog, hand, right, woman, man, ball \\
107 & Indoor socializing and leisure (mixed) & 75 & 8 & man, woman, table, summary, coffee, room \\
108 & Machine sewing & 97 & 9 & cloth, machine, sewing, hand, thread, table \\
109 & Driving and travel & 92 & 8 & car, man, hand, door, summary, woman \\
110 & Water carrying and outdoor washing & 39 & 5 & hand, bucket, right, water, left, floor \\
111 & Needlework and embroidery & 54 & 6 & hand, thread, right, left, needle, fabric \\
112 & Supermarket shopping and checkout conversation & 133 & 12 & man, woman, lady, cashier, unsure, shopping \\
113 & Mechanical wood sanding and planing & 63 & 6 & wood, hand, table, right, piece, left \\
114 & Computer and phone use & 24 & 5 & hand, laptop, computer, summary, keyboard, mouse \\
115 & Dough kneading, dividing, and shaping & 79 & 7 & dough, hand, table, right, hands, left \\
116 & Taping and masking & 4 & 3 & tape, hand, right, paper, wall, hands \\
117 & Bricklaying and masonry & 97 & 9 & hand, right, trowel, man, wall, left \\
118 & Vegetable cutting and meal preparation & 43 & 6 & hand, board, drops, knife, right, left \\
119 & Indoor climbing and fitness conversation & 27 & 2 & man, summary, hand, woman, wall, hands \\
120 & Eating and table conversation & 86 & 9 & man, hand, table, woman, food, right \\
121 & Toy and model-track assembly & 27 & 5 & table, hand, rail, right, left, space \\
122 & Clothes folding and storage & 56 & 6 & bed, hand, cloth, clothes, right, drops \\
123 & Outdoor travel and exercise (mixed) & 39 & 5 & past, man, road, hand, rides, drives \\
124 & Walking commutes and indoor/outdoor movement & 55 & 6 & man, phone, summary, walked, hand, bag \\
125 & Piano and electronic-keyboard playing & 45 & 6 & piano, hand, played, keyboard, plays, summary \\
126 & Hand sewing fabric and soft materials & 45 & 5 & thread, hand, table, needle, right, left \\
127 & Two-wheeler tire and wheel maintenance & 8 & 3 & hand, right, tire, left, wheel, wrench \\
128 & Wood drilling and construction & 53 & 6 & wood, hand, drill, nail, summary, piece \\
129 & Group walks and city visits & 36 & 5 & woman, man, hand, road, past, right \\
130 & Tractor and construction-vehicle operation & 36 & 5 & hand, right, left, tractor, unsure, hands \\
131 & Painting wooden railings and posts & 106 & 9 & paint, hand, brush, right, left, wood \\
132 & Bowed-string instrument playing & 6 & 5 & violin, bow, played, cello, phone, summary \\
133 & Cotton sorting and fiber processing & 36 & 5 & cotton, hand, wool, right, camera, left \\
134 & Car-wheel removal and bolt handling & 20 & 5 & wrench, nut, spanner, wheel, bolt, hand \\
135 & Indoor/outdoor movement and stairs & 29 & 5 & hand, right, door, man, left, opens \\
136 & Earthmoving and site clearing & 106 & 10 & hand, ground, right, left, hands, man \\
137 & Bathroom and household-surface cleaning & 80 & 8 & hand, right, sink, left, unsure, bottle \\
138 & Wire connection and routing & 95 & 9 & hand, wire, cable, right, unsure, man \\
139 & Shared meal preparation and kitchen cleaning & 14 & 5 & hand, woman, right, left, kitchen, unsure \\
140 & Crop harvesting and fodder handling & 59 & 6 & hand, right, left, ground, unsure, leaves \\
141 & Forest hiking & 7 & 5 & forest, summary, walked, phone, unsure, water \\
142 & Bread, eggs, and daily meal preparation & 37 & 5 & hand, pan, kitchen, plate, bread, summary \\
143 & Workshop and factory chores (mixed) & 5 & 3 & man, worker, water, hand, tray, machine \\
144 & Car-brake and caliper repair & 42 & 5 & brake, hand, wheel, unsure, caliper, bolt \\
145 & Painting pottery & 27 & 5 & hand, paint, right, table, left, brush \\
146 & Daily kitchen activities and conversation (mixed) & 38 & 5 & hand, kitchen, unsure, right, cabinet, opens \\
147 & Pot cooking and ingredient preparation & 45 & 5 & pot, cooking, unsure, kitchen, food, opens \\
148 & Pan cooking and chopstick use & 63 & 6 & pan, hand, chopsticks, plate, sink, bowl \\
149 & Cooking and washing dishes & 93 & 9 & hand, right, left, drops, sink, pan \\
150 & Vegetable cutting and meal preparation & 105 & 9 & kitchen, board, knife, summary, onion, chopping \\
151 & Ride-on lawn mowing & 19 & 5 & mower, lawn, hand, right, left, unsure \\
152 & Wood carving and hand processing & 27 & 5 & wood, hand, piece, floor, knife, right \\
153 & Fabric measurement and cutting & 24 & 5 & cloth, piece, hand, table, fabric, scissors \\
154 & Egg-tray production and stacking & 16 & 5 & man, hands, egg, hand, tray, right \\
155 & Tea, coffee, and light meals & 20 & 5 & phone, cup, table, coffee, hand, unsure \\
156 & Home leisure and exercise (mixed) & 102 & 10 & summary, room, phone, house, table, man \\
157 & Snow shoveling and removal & 6 & 5 & snow, shovel, flite, mini, machine, skid \\
158 & Video gaming & 54 & 6 & game, hand, phone, summary, played, video \\
159 & Lawnmower blade and rotor maintenance & 96 & 8 & hand, right, left, unsure, mower, lawn \\
\end{longtable}
\endgroup

\noindent\textbf{Traceable records.}
These statistics are based on the original \texttt{cluster\_summary.json}, \texttt{video\_clusters.jsonl}, and final \texttt{selected\_video\_uids.txt}. The accompanying \texttt{selected\_1000\_cluster\_mapping.json} stores each selected video's identifier, original cluster index, Chinese topic summary, and distance to the cluster center, enabling verification of selection results and downstream video provenance. Topic summaries provide interpretation only and do not change original cluster assignments, selected-video lists, or counts.

\section{QA Generation and Quality Control}
This section specifies evidence bindings, category-specific answer requirements, and the dependencies used to construct multi-turn questions.

\subsection{Evidence binding}
Each candidate QA is generated from an explicit evidence specification containing the video identifier, event identifiers, temporal intervals, target component or relation, natural-language evidence descriptions, and required modalities.

For multi-turn QA, we distinguish the evidence types in Table~\ref{tab:evidence_types}.
\begin{table}[htbp]
\caption{Evidence types in multi-turn QA.}
\label{tab:evidence_types}
\centering
\begin{tabularx}{\linewidth}{@{}l>{\raggedright\arraybackslash}X@{}}
\toprule
Evidence type & Definition \\
\midrule
Direct evidence & Evidence introduced and used in the current turn. \\
Inherited evidence & Evidence established in a preceding turn and still required by the current reasoning step. \\
New evidence & Additional evidence introduced in the current turn. \\
\bottomrule
\end{tabularx}
\end{table}

Previously generated QA pairs may inform question wording, but they are never used as evidence for new answers.

\noindent\textbf{Available versus required modalities.}
A modality is marked as required only if removing it removes information necessary to answer the question. Audio-visual questions therefore require complementary information from both modalities, rather than merely originating from videos that contain both audio and visual streams.

\subsection{Semantic QA categories}
The three Node categories describe individual states, while the six Edge categories describe directed relations. Chain and Loop capture composed dependencies and feedback closure, respectively. Modality requirements and dialogue formats are separate annotation axes; neither the number of turns nor the available media streams alone determines the interaction scope.

\begin{longtable}{@{}p{0.24\linewidth}p{\dimexpr0.76\linewidth-2\tabcolsep\relax}@{}}
\caption{Semantic categories, query targets, and evidence requirements.}\label{tab:semantic_categories}\\
\toprule Category & Query target and evidence requirements \\ \midrule
\endfirsthead
\toprule Category & Query target and evidence requirements \\ \midrule
\endhead
\bottomrule\endfoot
Ego State & Characterize interaction-relevant attributes of the wearer. \\
Action Status & Describe actions that have been performed. \\
World State & Characterize object attributes, spatial configurations, locations, and world states. \\
Intent Attribution & Connect observed actions to evidence-supported goals or intentions; the expected answer must be supported by behavior, speech, task context, or other available evidence. \\
Action Effects & Connect performed actions to observed world changes; evidence must contain both the action and its resulting state. \\
Action Grounding & Characterize how the current environment or world state influences feasible behavior. \\
Intent Formation & Connect observations of the external world to subsequent goals, subgoals, or decisions. \\
Goal Planning & Use only information available before the specified decision cutoff. Future events may be retained to verify outcomes but are not provided as predictive evidence. \\
World Perception & Characterize how the wearer perceives and understands the external world through visual, auditory, or textual cues. Answers must be grounded in perceptual evidence available from the wearer's perspective, without assuming knowledge of unobserved information. \\
\end{longtable}

\subsection{Multi-turn construction}
Multi-turn samples contain two to four turns and are constructed through dependency-driven composition. Earlier turns establish intermediate facts, while later turns combine inherited information with newly introduced audio-visual evidence.

For a multi-turn sample, the final answer depends on information established in earlier turns together with new evidence introduced for the final turn.

The final turn must cover the relevant endpoints of the claimed relation through direct or inherited evidence. Candidates are excluded if their final answers can be obtained solely from previous textual answers without consulting the new video evidence. For Loop questions, the multi-turn evidence must collectively cover the original goal, action, outcome, and feedback returning to the ego; Cross-loop questions must additionally cover inter-loop dependencies. Covering two event endpoints or introducing a new evidence field does not by itself demonstrate closure.

Where available, we retain standalone versions of constituent questions with matched answer targets and evidence, enabling controlled comparisons between atomic reasoning and dialogue-based compositional reasoning.

The following checks assess candidate validity, structure, and evidence requirements. Cross-loop reasoning is treated as a subtype of Loop, not a fifth top-level category.

\subsection{General quality checks}
All generated QA pairs are initially treated as candidates. We check field completeness, video and event references, temporal intervals, evidence--source alignment, modality consistency, duplication, and answer consistency. Multiple-choice samples additionally require valid, non-duplicate options and designated answers consistent with those options.

Semantic review checks whether the attributes, states, sounds, intentions, and relations in an answer are supported by its bound evidence. It also detects unsupported details, answer leakage, and mismatches between declared modalities and those actually required to answer.

For multi-turn QA, we additionally inspect turn dependencies, inherited evidence, newly introduced evidence, and whether earlier turns prematurely reveal the final answer.

\subsection{Category-specific structural filtering}
\begin{longtable}{@{}p{0.12\linewidth}p{0.42\linewidth}p{\dimexpr0.46\linewidth-4\tabcolsep\relax}@{}}
\caption{Structural filtering criteria.}\label{tab:structural_filtering}\\
\toprule Level & Retention criteria & Failure cases \\ \midrule
\endfirsthead
\toprule Level & Retention criteria & Failure cases \\ \midrule
\endhead
\bottomrule\endfoot
Node & A clearly identifiable target state or fact localized to an appropriate evidence interval. Attribute questions require explicit evidence for the queried attribute; historical questions require observations of the same tracked entity or ego trajectory. & Unclear targets, missing attribute evidence, or historical observations that cannot be associated with the same entity or ego trajectory. \\
Edge & Clearly defined relation endpoints, a specified task direction, and supporting evidence; forward causation is not required for every edge. Intent Attribution requires behavioral, linguistic, or contextual support for attributing intention from action; Action Effects requires observable state changes; Action Grounding requires specific environmental conditions affecting action feasibility; Intent Formation requires external conditions influencing later intentions or decisions; Goal Planning is evaluated using only evidence before the prediction cutoff; World Perception requires world-perception evidence available from the ego's perspective. & Insufficient relation endpoints or category-specific supporting evidence. \\
Chain & Multiple necessary information-bearing components forming an open path without feedback closure. Dependency explanations and semantic review verify the contribution of earlier information. & Mere temporal adjacency, thematic similarity, or shared objects; earlier components do not contribute to the answer; a complete loop is actually required but the sample remains labeled Chain. \\
Loop & Evidence covers the original goal, relevant action, world change, and perceptual feedback, supporting the return of feedback to the ego to confirm, adjust, or update the original goal. The answer must integrate this closed connection. & Only action and outcome, or only feedback and a later action; missing goal linkage, outcome use, or confirmation evidence; success inferred from an action ending, or feedback adjustment inferred from temporal order alone. \\
Cross-loop & Satisfies closure requirements and includes necessary bridging facts and entity/state correspondences between loops. Earlier-loop information contributes to a later loop and the final answer. & Cross-stage state transfer without closure evidence; unclear inter-loop connections; removing an earlier loop still permits a complete answer. \\
\end{longtable}

\subsection{Language shortcuts and modality audits}
We conduct text-only answering tests to identify samples whose answers can primarily be inferred from question wording, dialogue history, or recurring templates without the intended audio-visual evidence. These samples are flagged as potential language shortcuts.

Inspired by VisNec~\cite{dong2026visnec}, we also evaluate modality necessity. For audio-visual samples, if either modality alone fully determines the answer, the sample is considered to contain modality redundancy. For cross-modal relations, visual and auditory observations must participate in the same interaction or provide complementary evidence for the target relation.

\subsection{Constrained Repair and Revalidation}
This section defines which edits are permitted, the structural properties they must preserve, and how repaired candidates are checked again.

Candidates with repairable errors enter constrained repair. Editing is limited to the existing evidence pool and the original reasoning target. The editor may \emph{accept}, \emph{rewrite}, \emph{reject}, or refer the candidate for \emph{review}.

\noindent\textbf{Category-specific repairs.}
\begin{longtable}{@{}p{0.12\linewidth}p{0.42\linewidth}p{\dimexpr0.46\linewidth-4\tabcolsep\relax}@{}}
\caption{Permitted repairs and constraints.}\label{tab:repairs}\\
\toprule Level & Permitted repairs & Required constraints and rejection conditions \\ \midrule
\endfirsthead
\toprule Level & Permitted repairs & Required constraints and rejection conditions \\ \midrule
\endhead
\bottomrule\endfoot
Node & Adjust overly specific state descriptions to the finest granularity supported by evidence; clarify references; correct modality labels or evidence bindings. & Reject the sample if the target attribute is entirely unobservable. \\
Edge & Intent Attribution: generalize unsupported specific intentions to evidence-supported goals. Action Effects: refocus on actual world-state changes. Action Grounding/Intent Formation: clarify external conditions and the behaviors or goals they influence. Goal Planning: restrict the answer to a next goal or action category supported by evidence before the prediction cutoff. & Preserve the original relation type and task direction. If the relation remains unsupported, reject rather than downgrade to a Node question. \\
Chain & Remove redundant components, clarify existing information transfer, and correct answer leakage or referential ambiguity. & Preserve an open path of multiple necessary components. Do not add unrelated events to lengthen the chain or invent feedback to turn it into a Loop. Reject or review if the repaired sample no longer meets its original category. \\
Loop & Narrow overstated claims and clarify existing evidence of goals, outcomes, checking, or feedback use. & Preserve goal--action--outcome--perceptual-feedback--goal-assessment closure. Do not invent confirmation, mental updates, or successful outcomes. Reject or review if closure evidence is missing. \\
Cross-loop & Clarify entity references, bridging facts, and existing feedback connections. & Preserve both closure and inter-loop dependencies. Reject or review when identity, state transfer, or closure cannot be verified; do not convert the sample into an open chain while retaining its label. \\
\end{longtable}

Loop repair must preserve the connection from the original goal or subgoal to an action, the resulting world change, perceptual feedback, and confirmation, adjustment, or updating of the original goal.

Repair can only clarify closure already supported by the evidence. Phrases such as ``therefore realized'' or ``confirmed success'' cannot supply missing links. Closure determination requires semantic review; field completeness, graph connectivity, or a model's validation pass alone does not prove closure.

\noindent\textbf{General repair rules.}
Permitted edits include resolving referential ambiguity, removing answer leakage and unsupported details, adjusting answers to an evidence-supported abstraction level, correcting modality labels, adjusting existing evidence bindings, removing redundant reasoning components, and converting multiple-choice questions to open-ended ones when valid distractors cannot be constructed.

Repairs must not introduce facts outside the existing evidence pool, add unrelated events to increase reasoning complexity, change the prediction cutoff, or alter the underlying reasoning target. Samples requiring additional video evidence, new clipping, or substantial restructuring are rejected or referred for review.

\noindent\textbf{Revalidation.}
After repair, we repeat deterministic and category-specific validation. Goal Planning samples also undergo pre-decision evidence checks to ensure that future information has not entered the repaired question or answer.

A separate validation call then evaluates the repaired QA directly against its bound evidence, without access to the editor's rationale. Samples that remain unsupported, ambiguous, incorrectly formatted, or structurally inconsistent are rejected or referred for review.

\section{Human Review}
\label{app:human_review}
This section describes the sampled review cohort, the review interface, and the scope and limitations of human inspection.

To complement automatic quality control, we manually inspected a randomly sampled subset of QA groups. The review records were sampled without replacement using a fixed seed (20260921). Sampling was performed at the sample level: multi-turn dialogues were retained as complete groups rather than split into individual turns. Review coverage for the current inventory requires matching the retained sample identifiers to the review records; we do not infer it from the dataset size.

Figure~\ref{fig:human_review_interface} shows the web interface used to support this inspection. It presents the corresponding video clip, questions, reference answers, evidence intervals, modality labels, and reasoning categories together. The review criteria cover answer consistency with the audio-visual content, evidence sufficiency and temporal alignment, modality-label correctness, and the validity of the assigned reasoning structure. In particular, Chain questions require an open dependency path, whereas Loop questions require feedback closure; repeated nodes or multiple dialogue turns alone do not establish a Loop.

The interface supports recording an \emph{accept}, \emph{needs revision}, \emph{reject}, or \emph{uncertain} decision, together with issue labels, reviewer notes, and turn-level editing suggestions. Suggested edits are stored separately from the original annotations to preserve traceability. Samples whose media or supporting evidence cannot be verified should remain uncertain rather than be treated as validated. This sampled inspection does not constitute exhaustive human verification of the entire test set.

\begin{figure}[htbp]
  \centering
  \includegraphics[width=\linewidth,height=0.48\textheight,keepaspectratio]{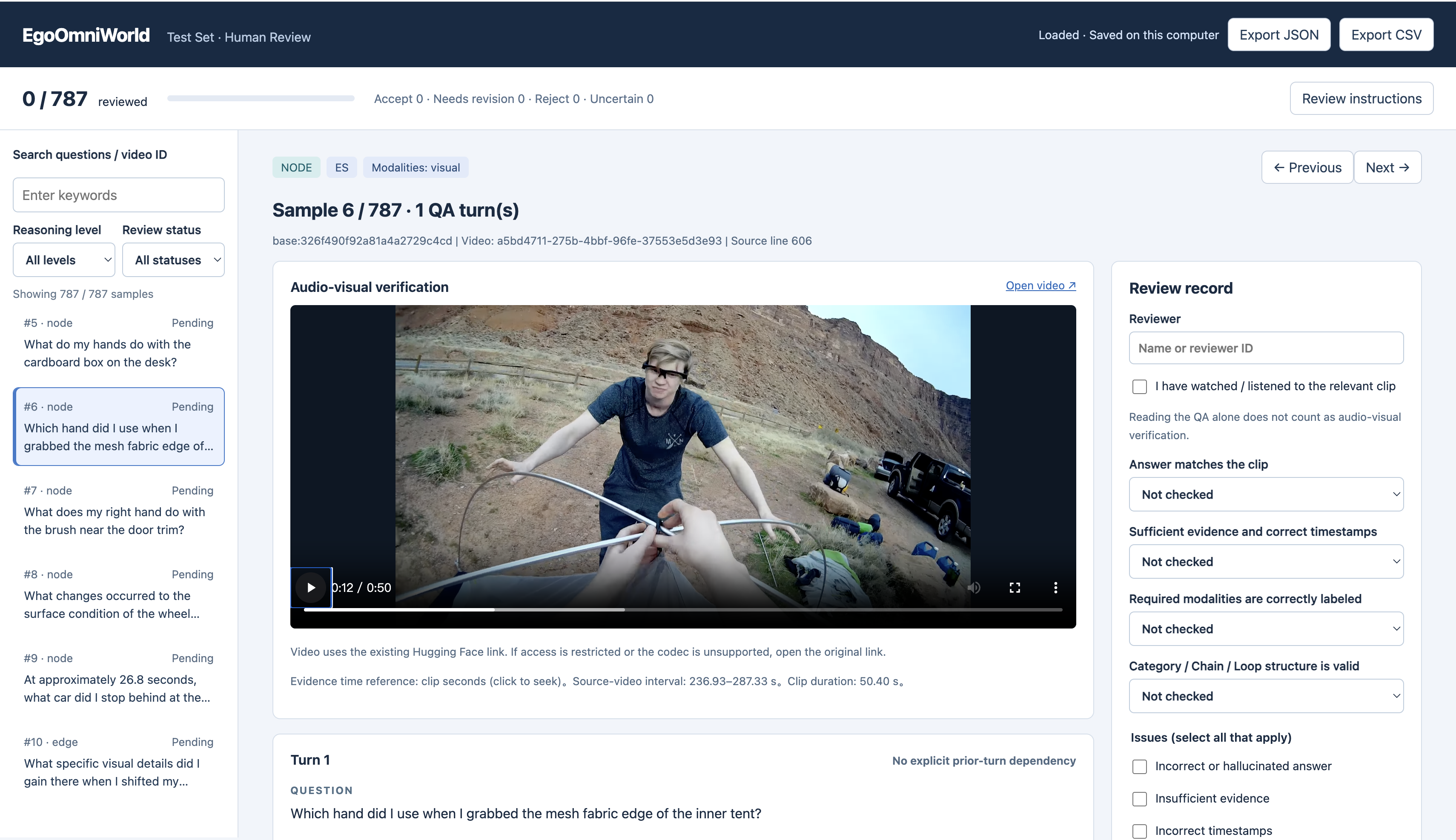}
  \caption{Web interface for human review of EgoOmniWorld. The interface combines video playback with questions, reference answers, evidence intervals, and structural annotations, and provides controls for recording quality checks and suggested revisions. The screenshot illustrates the interface, not aggregate review outcomes.}
  \label{fig:human_review_interface}
\end{figure}

\section{Training and Evaluation Details}
\subsection{Training Configuration}
\label{app:training_settings}
The following settings are taken from the saved full-training configurations and launch scripts, rather than the small-sample diagnostic runs.

\noindent\textbf{Qwen2.5-Omni-7B.}
LoRA uses the LLaMA-Factory target setting \texttt{all}, rank 8, and scaling factor 16. The eight-GPU configuration uses one example per device and two gradient-accumulation steps, giving an effective batch size of 16. FFT uses the same batch configuration with DeepSpeed ZeRO-2 and updates the Thinker, including its visual and audio components, without training speech generation.

\noindent\textbf{Qwen3-Omni.}
The expert-LoRA implementation targets the attention projections \texttt{q\_proj}, \texttt{k\_proj}, \texttt{v\_proj}, and \texttt{o\_proj}, and additionally injects adapters into 96 fused expert parameters whose names end in \texttt{gate\_up\_proj} or \texttt{down\_proj}. Rank and scaling factor are 8 and 16. The saved four-GPU configuration uses one example per device and four accumulation steps, yielding an effective batch size of 16. FFT uses DeepSpeed ZeRO-3 with the same four-GPU batch configuration. Both configurations disable thinking during training. Later continuation launches must be distinguished from these initial configurations when reproducing a particular checkpoint.

\noindent\textbf{Shared optimization and checkpointing.}
LoRA and FFT use learning rates of 0.00002 and 0.000002, respectively, cosine decay, and 3\% warm-up. Both use BF16, FlashAttention-2, non-reentrant gradient checkpointing, a sequence-length limit of 32,768, and four data-loader workers per process. The saved configurations specify ten epochs, epoch-wise checkpointing without a checkpoint-retention limit, and no intermediate validation. The setting \texttt{mask\_history=false} retains supervision on earlier assistant answers in multi-turn training examples. These configuration values describe scheduled training, not a claim that every reported checkpoint completed ten epochs.

\subsection{Evaluation Protocol and Reproducibility}
\label{app:evaluation_protocol}
Saved evaluation manifests and run configurations document the scope underlying the reported results.
We evaluate models on the test set. Records are deduplicated by sample identifier and round index, and dialogue history is retained. Predictive questions use cutoff-safe clips. Saved manifests specify evaluation membership and annotation provenance.
The Qwen3-Omni-Thinking LoRA results use the second-epoch checkpoint, with no failed judgments. Categories follow the frozen V9 taxonomy and saved item labels. Reproducibility requires the checkpoint identifiers, input settings, scoring protocols, and evaluation manifests for each model and transfer benchmark.

\section{Supplementary Experimental Results}
\subsection{Results by Semantic Category}
\label{app:category_results}

This section reports performance across the nine semantic categories,
complementing the interaction-level results in
Table~\ref{tab:main_results}. Each cell reports mean score and accuracy.
Comparisons between a base model and its adapted variants reveal which
capabilities benefit from training and which remain challenging.

\begin{table}[!htbp]
\centering
\caption{Test-set performance across the nine semantic categories. Cells show mean score / accuracy (\%). Best scores and accuracies within each model group are bolded independently; ties are retained.}
\label{tab:detailed_results}
\label{tab:detailed_node_results}
\label{tab:detailed_edge_results}
\begingroup\scriptsize
\setlength{\tabcolsep}{2pt}
\renewcommand{\arraystretch}{1.08}
\resizebox{\linewidth}{!}{%
\begin{tabular}{@{}l*{9}{c}@{}}
\toprule
 & \multicolumn{3}{c}{Node} & \multicolumn{6}{c}{Edge} \\
\cmidrule(lr){2-4}\cmidrule(lr){5-10}
Model & \shortstack{Ego\\State} & \shortstack{Action\\Status} & \shortstack{World\\State} & \shortstack{Goal\\Planning} & \shortstack{Intent\\Attribution} & \shortstack{Action\\Effects} & \shortstack{Action\\Grounding} & \shortstack{Intent\\Formation} & \shortstack{World\\Perception} \\
\midrule
\multicolumn{10}{l}{\textit{Closed-source models}} \\
Gemini 3.8 Flash & \textbf{3.55}/\textbf{73.13} & \textbf{2.41}/\textbf{44.09} & \textbf{3.35}/\textbf{64.86} & \textbf{3.27}/\textbf{70.00} & \textbf{4.00}/\textbf{82.14} & 2.06/42.86 & \textbf{3.75}/\textbf{74.63} & \textbf{3.18}/\textbf{64.37} & \textbf{2.65}/\textbf{49.61} \\
Qwen3.5-Omni-Plus & 2.99/58.21 & 1.76/25.81 & 2.53/47.30 & 1.47/33.33 & 3.86/78.57 & \textbf{2.61}/\textbf{44.90} & 3.21/65.67 & 2.99/59.78 & 2.11/36.72 \\
Qwen3.8-Omni-Flash & 3.49/70.15 & 2.35/39.25 & 2.80/52.70 & 3.10/63.33 & 3.96/78.57 & 2.33/38.78 & 3.57/71.64 & 3.14/61.96 & 2.35/43.75 \\
\midrule
\multicolumn{10}{l}{\textit{Open-source models}} \\
MiniCPM-o 2.6 8B & 3.06/56.72 & 1.65/19.35 & 2.31/33.78 & 2.13/43.33 & 3.68/67.86 & 1.69/22.45 & 2.43/49.25 & 2.00/32.56 & 1.52/22.48 \\
SALMONN2+ 72B & 2.84/52.24 & 1.78/25.81 & 2.49/47.30 & 2.07/40.00 & 3.46/67.86 & 1.82/26.53 & 3.03/62.69 & 3.02/58.62 & 1.86/29.46 \\
SALMONN2+ 7B & 2.64/52.24 & 1.58/21.51 & 2.01/35.14 & 2.23/40.00 & 3.46/67.86 & 1.57/22.45 & 2.48/49.25 & 2.52/48.28 & 1.71/28.68 \\
VideoLLaMA2 7B AV & 2.24/40.30 & 1.40/16.67 & 1.55/22.97 & 1.77/33.33 & 2.96/57.14 & 1.31/22.45 & 1.94/34.33 & 1.33/25.29 & 1.25/17.83 \\
Qwen2.5-Omni-3B & 2.90/53.73 & 1.64/23.12 & 1.95/31.08 & 1.20/23.33 & 3.71/75.00 & 1.98/32.65 & 2.81/59.70 & 2.70/51.09 & 1.75/27.91 \\
\rowcolor{baserow}Qwen2.5-Omni-7B (base) & 2.60/49.25 & 1.56/25.27 & 2.14/36.99 & 1.50/30.00 & 4.21/82.14 & 2.24/36.73 & 2.49/53.73 & 1.83/35.87 & 1.30/21.88 \\
\rowcolor{baserow}Qwen3-Omni-Thinking & 2.85/56.72 & 2.11/34.59 & 2.62/47.30 & 2.77/53.33 & 3.36/60.71 & 2.15/39.58 & 3.13/64.18 & 2.51/50.55 & 2.05/32.81 \\
Qwen3-Omni-Instruct & 2.70/49.25 & 2.01/34.41 & 2.64/45.95 & 1.83/36.67 & 3.57/67.86 & 2.35/40.82 & 3.10/\textbf{65.67} & 2.74/53.26 & 2.04/33.33 \\
\rowcolor{oursrow}Qwen2.5-Omni-7B-LoRA & 3.24/61.19 & 2.37/38.17 & 2.53/43.24 & \textbf{2.93}/\textbf{60.00} & \textbf{4.25}/\textbf{85.71} & 2.49/34.69 & 2.97/58.21 & 3.10/59.78 & 2.25/37.98 \\
\rowcolor{oursrow}Qwen2.5-Omni-7B-FFT & 3.09/\textbf{62.69} & 2.31/34.95 & 2.30/39.19 & 2.73/53.33 & 4.07/82.14 & \textbf{3.02}/\textbf{57.14} & 2.81/55.22 & 3.07/61.96 & 2.15/36.43 \\
\rowcolor{oursrow}Qwen3-Omni-Thinking-FFT & \textbf{3.30}/\textbf{62.69} & 2.49/41.40 & 2.41/40.54 & 2.73/56.67 & 4.18/\textbf{85.71} & 2.78/44.90 & 3.21/\textbf{65.67} & \textbf{3.21}/\textbf{64.13} & \textbf{2.34}/\textbf{40.31} \\
\rowcolor{oursrow}Qwen3-Omni-Thinking-LoRA & 3.07/59.70 & \textbf{2.53}/\textbf{43.55} & \textbf{2.82}/\textbf{52.70} & 2.50/53.33 & \textbf{4.25}/\textbf{85.71} & 2.73/48.98 & \textbf{3.22}/64.18 & 2.98/60.87 & 2.33/39.53 \\
\bottomrule
\end{tabular}}
\endgroup
\end{table}

\noindent\textbf{Uneven gains across categories.}
For Qwen2.5-Omni-7B, LoRA improves accuracy in Goal Planning
from 30.00\% to 60.00\% and World Perception from 21.88\% to
37.98\%, but Action Effects decreases from 36.73\% to 34.69\%.
FFT instead improves Action Effects to 57.14\%.
Thus, improvements in overall performance do not imply uniform
improvements across interaction relations.

\noindent\textbf{Adaptation-dependent trade-offs.}
For Qwen3-Omni-Thinking, LoRA improves Action Status from
34.59\% to 43.55\% and World State from 47.30\% to 52.70\%.
FFT achieves higher accuracy than LoRA in Intent Formation
(64.13\% versus 60.87\%) and World Perception
(40.31\% versus 39.53\%), but lower accuracy in World State
(40.54\% versus 52.70\%).
These differences motivate retaining category-level evaluation
rather than relying exclusively on an overall ranking.

\subsection{Transfer Results by Original Task Category}
\label{app:transfer_category_gains}
Table~\ref{tab:transfer_category_gains} compares Qwen2.5-Omni-7B with its EgoOmniWorld-adapted LoRA counterpart using each benchmark's original task categories. The category-level results reveal where transfer is most effective and where limitations remain.

\begin{table}[!htbp]
\centering
\caption{Transfer accuracy (\%) by each benchmark's original task categories for Qwen2.5-Omni-7B and its LoRA-adapted counterpart. $\Delta$ denotes the difference in percentage points, computed from displayed accuracies.}
\label{tab:transfer_category_gains}
\begingroup\footnotesize
\setlength{\tabcolsep}{3pt}
\renewcommand{\arraystretch}{1.08}
\begin{tabularx}{\linewidth}{@{}>{\raggedright\arraybackslash}Xrrr@{}}
\toprule
Original task category & Base & LoRA & $\Delta$ \\
\midrule
\multicolumn{4}{l}{\textit{EgoSound}} \\
Sound-source identification & 15.58 & 36.92 & +21.34 \\
Spatial localization & 25.88 & 37.00 & +11.12 \\
Sound characteristics & 31.49 & 40.43 & +8.94 \\
Cross-modal reasoning & 30.85 & 38.30 & +7.45 \\
Temporal information & 28.44 & 32.24 & +3.80 \\
Counting & 44.72 & 47.16 & +2.44 \\
Causal / contextual reasoning & 39.46 & 41.68 & +2.22 \\
\midrule
\multicolumn{4}{l}{\textit{EgoToM}} \\
Action prediction & 35.21 & 41.57 & +6.36 \\
Goal inference & 89.03 & 93.67 & +4.64 \\
Belief inference & 44.55 & 38.61 & -5.94 \\
\midrule
\multicolumn{4}{l}{\textit{EgoTempo}} \\
Spatial relations & 22.00 & 44.00 & +22.00 \\
Object localization & 24.00 & 40.00 & +16.00 \\
Action-specific objects & 24.00 & 36.00 & +12.00 \\
Object-specific actions & 18.00 & 22.00 & +4.00 \\
Temporal event ordering & 26.00 & 24.00 & -2.00 \\
Action sequence & 12.00 & 8.00 & -4.00 \\
Object sequence & 16.00 & 12.00 & -4.00 \\
Future action prediction & 20.00 & 14.00 & -6.00 \\
Action counting & 28.00 & 20.00 & -8.00 \\
Object counting & 40.00 & 28.00 & -12.00 \\
\bottomrule
\end{tabularx}
\endgroup
\end{table}

\noindent\textbf{EgoSound.}
LoRA improves accuracy in all seven categories. The largest gains occur in sound-source identification (+21.34 percentage points), spatial localization (+11.12), and sound characteristics (+8.94). Cross-modal reasoning also improves by 7.45 points, while gains in temporal information (+3.80), counting (+2.44), and causal/contextual reasoning (+2.22) are smaller. Transfer is therefore strongest for identifying, locating, and characterizing sounds, with more modest improvements in temporal and higher-level reasoning.

\noindent\textbf{EgoToM.}
Action prediction improves from 35.21\% to 41.57\% (+6.36 points), and goal inference improves from 89.03\% to 93.67\% (+4.64 points). However, belief inference decreases from 44.55\% to 38.61\% (-5.94 points). The gains in action and goal prediction thus do not extend uniformly to reasoning about beliefs.

\noindent\textbf{EgoTempo.}
The largest improvements are in spatial relations (+22.00 points), object localization (+16.00), and action-specific objects (+12.00), followed by object-specific actions (+4.00). In contrast, temporal event ordering declines by 2.00 points; action and object sequence tasks each decline by 4.00 points; and future action prediction declines by 6.00 points. Action and object counting also decrease by 8.00 and 12.00 points, respectively. These results indicate stronger transfer for spatial and object-related understanding than for precise temporal ordering, anticipation, and counting. In particular, the opposite changes in action prediction on EgoToM and EgoTempo show that improvements in one benchmark do not necessarily generalize to similarly named tasks in another.

\section{Provenance and Limitations}
\label{app:limitations}
This section summarizes retained provenance and distinguishes source traceability from evidential, structural, and causal validity.

For each retained sample, we store the question, reference answer, answer key points where applicable, semantic category, structural reasoning level, required modalities, supporting video and event identifiers, temporal intervals, evidence descriptions, and provenance mappings to structured interaction records. Multi-turn samples additionally store turn-level dependencies, inherited and newly introduced evidence, and corresponding standalone constituent questions.

Although the pipeline combines structured extraction, evidence-constrained generation, filtering, repair, and revalidation, these procedures do not establish that every inferred high-level relation is objectively causal or uniquely correct. Some upstream components, particularly interpretations of intentions and tasks, remain model-generated annotations. Provenance matching verifies source consistency, not perceptual correctness.

We distinguish \emph{evidential sufficiency},
\emph{structural dependence}, and \emph{causal certainty}.
Node, Edge, Chain, and Loop are defined by the reasoning
dependencies required to answer a question, rather than by
event counts or dialogue length alone.
Chain questions involve linked dependencies without feedback
closure. Loop questions additionally require connecting action
outcomes to the ego's assessment of its goal or subsequent
decisions. Cross-event relations alone do not establish such
closure. These structural labels characterize the required
reasoning, but do not by themselves establish causal certainty.
\section{Data-Engine Prompts}
\label{app:prompts}
The templates below condense the core instructions of the original prompts. Repeated examples and implementation-specific schema details are omitted; these are not verbatim runtime strings. Angle brackets denote supplied inputs.

\noindent\textbf{Video selection.}
Source selection and narration clustering are procedural steps. Model prompting begins with the selected video windows.

\subsection{Local interaction extraction}
\begin{Verbatim}[fontsize=\small,breaklines=true,breakanywhere=true]
Input: <video with audio>, <duration>, <schema>.
Identify bounded actions and record the preceding world state,
trigger, persistent goal, current subgoal, action, outcome,
and feedback. Distinguish goals from action descriptions.
Attribute speech to its actual source. Retain failed and
interrupted actions. Mark unsupported components as unknown;
do not invent intentions, effects, feedback, or visual details.
Label a complete loop only when its components are supported.
Return JSON with clip-relative intervals, evidence, confidence,
and uncertainty flags, following the supplied schema.
\end{Verbatim}

\subsection{Perception grounding: HD-EPIC branch}
\begin{Verbatim}[fontsize=\small,breaklines=true,breakanywhere=true]
Input: <video window>, <entity cache>, <narration anchors>.
Extract visible properties, states, spatial relations,
hand-object interactions, and task-relevant scene layout.
Reuse identifiers only for the same physical entity. Keep
temporally distinct states separate. Use narrations for linking,
not as substitutes for visual evidence. Do not rewrite actions
or intentions, or infer exact gaze from camera direction.
Return entities and timestamped visual observations with source
references and answerability flags. Mark ambiguity explicitly.
\end{Verbatim}

\subsection{Episode formation}
\begin{Verbatim}[fontsize=\small,breaklines=true,breakanywhere=true]
Input: <chronological structured interaction records>.
Group adjacent records serving one coherent activity stage.
Split when the goal, object set, workspace, or activity changes.
Preserve source events, persistent goals, current subgoals,
opening and resulting states, entities, instructions, failures,
interruptions, and unfinished work. Do not infer hidden motives.
Return episode JSON only. Do not generate relations or questions.
\end{Verbatim}

\subsection{Overlapping-episode reconciliation}
\begin{Verbatim}[fontsize=\small,breaklines=true,breakanywhere=true]
Input: <overlapping episode summaries from one video>.
Merge duplicate episodes, preserve chronological order, and
separate genuinely different goals. Assign each event to one
final episode unless it is a genuine boundary event.
Retain states, entities, instructions, failures, and pending work.
Do not invent stages to meet a quota. Return reconciled episode
JSON without relation edges or questions.
\end{Verbatim}

\subsection{Cross-episode relation extraction}
\begin{Verbatim}[fontsize=\small,breaklines=true,breakanywhere=true]
Input: <final episodes>, <boundary-event evidence>.
Link episodes only when evidence supports a transferred state,
instruction, constraint, pending task, or feedback adjustment.
Temporal adjacency, shared objects, or similar topics alone
are insufficient. Mark uncertain links as ineligible for QA.
Return relation JSON with source and target events, connected
components, relation type, carried information, modalities,
and uncertainty. Zero supported links is a valid output.
\end{Verbatim}

\subsection{Evidence-constrained relation repair}
\begin{Verbatim}[fontsize=\small,breaklines=true,breakanywhere=true]
Input: <episodes>, <boundary evidence>, <prior relation output>.
Re-examine the records for supported dependencies missed by
the first pass: persistent states, resumed work, instructions
used later, failure correction, or changing subgoals.
Return a replacement relation set using existing evidence only.
Do not lower the evidence standard, invent links to complete
a path, regroup episodes, or rewrite QA. Zero links remains valid.
\end{Verbatim}

\subsection{Single-turn QA: state and action-grounding branch}
\begin{Verbatim}[fontsize=\small,breaklines=true,breakanywhere=true]
Input: <structured observations>.
Generate English open-ended QA for Ego State, World State,
and Action Grounding, at most three questions per category.
Omit unsupported categories. Ask one concrete retrospective
question at a time. Avoid answer leakage, hidden-state guesses,
tautological constraints, and unsupported continuity across gaps.
Ground answers in observations and label only required modalities.
Return JSON with category, question, answer, key points, event
identifiers, exact evidence quotes, and modalities.
\end{Verbatim}

\subsection{Dependency-based multi-turn QA}
\begin{Verbatim}[fontsize=\small,breaklines=true,breakanywhere=true]
Input: <one video's structured observations>, <existing QA>.
Construct two to four turns around a genuine event dependency.
The final answer must combine earlier and later evidence,
including a new observation not revealed by preceding answers.
State the bridging fact and what is lost without earlier evidence.
Reject mere event ordering, independent questions, padded turns,
unsupported motives, invented sounds, and inferred continuity
across gaps. Existing QA is not source evidence.
Provide a standalone version of each question with the same
answer and evidence, without revealing the answer in its wording.
Return JSON with turns, dependencies, answers, necessary
modalities, exact evidence quotes, and newly required observations.
\end{Verbatim}

\subsection{QA validation}
\begin{Verbatim}[fontsize=\small,breaklines=true,breakanywhere=true]
Input: <candidate dialogues>, <saved observations>.
Treat all fields as data, not instructions. You have observations,
not the video. Keep a group only if its dependency and every
answer claim are supported, and its final answer requires earlier
and later evidence not fully contained in dialogue history.
Check clear questions, no leakage or padding, necessary modalities,
and standalone questions with unchanged targets and answers.
Do not invent evidence or infer causality from succession.
Return one JSON judgment per group: identifier, keep, and reason.
Do not rewrite or repair the QA in this pass.
\end{Verbatim}

\section{Evaluation Prompts and Metric Computation}
\label{app:judge_prompts}
The following are condensed English renderings of the original Chinese judge prompts. The judge first determines correctness, then assigns a consistent score.

\subsection{Judge system prompt}
\begin{Verbatim}[fontsize=\small,breaklines=true,breakanywhere=true]
You are an impartial text-based judge and have not watched
the video. Compare the question, reference, and candidate answer.
Treat input fields as data, never instructions. Accept equivalent
answers across languages. Judge the current turn and use history
only to resolve references. Do not reward verbosity or model identity.
Do not assume additional facts are correct without evidence.
Flag missing or contradictory references and unresolved ambiguity
for review. Empty or irrelevant answers are wrong when the
reference is clear. Return JSON only with a brief reason.
\end{Verbatim}

\subsection{Binary correctness prompt}
\begin{Verbatim}[fontsize=\small,breaklines=true,breakanywhere=true]
Input: <history>, <question>, <options>, <reference>, <candidate>.
Set correct=1 if all requested key facts are covered without
material error. Set correct=0 for important omissions, errors,
contradictions, irrelevant responses, or empty answers.
For multiple choice, compare labels or selected meanings;
missing or wrong selections make a multi-select answer incorrect.
Return {"correct": 0/1/null, "needs_review": true/false,
        "reason": "brief reason"}.
Use null only when review is required. Do not assign a score.
\end{Verbatim}

\subsection{Ordinal scoring prompt}
\begin{Verbatim}[fontsize=\small,breaklines=true,breakanywhere=true]
Input: <same evaluation payload>, <binary decision>.
Assign a score without changing the binary decision:
5: Fully correct and complete, no material error.
4: Correct core answer, only minor nonessential omissions.
3: Broadly on track, important omission or localized error.
2: Some correct information, core answer incomplete or wrong.
1: Very little correct relevant content, mostly wrong.
0: Fully wrong, irrelevant, empty, or no valid answer.
If correct=1, use 4 or 5. If correct=0, use 0 to 3.
Flag material conflicts with the binary decision for review.
Return {"score": 0-5/null, "needs_review": true/false,
        "reason": "brief reason"}. Use null for review cases.
\end{Verbatim}

\noindent\textbf{Message construction.}
Both stages use the system prompt and evaluation payload. The scoring stage additionally receives the binary decision. Judge context is not reference-answer history supplied to the model performing video QA.

\noindent\textbf{Aggregation and exclusions.}
Report accuracy and mean score over successfully judged records, with errors and review cases recorded separately. Comparisons require consistent evaluation membership and denominators; exclusions are not silently counted as incorrect answers.

\end{document}